\documentclass[twoside,11pt]{article}
\usepackage[preprint]{jmlr2e}
\usepackage{amsmath}
\usepackage{thm-restate}
\usepackage{booktabs}
\usepackage{makecell}
\usepackage{lastpage}
\jmlrheading{}{2026}{1-\pageref{LastPage}}{}{}
{Andrey A. Dukhovny and Andrey M. Lange}
{Andrey A. Dukhovny and Andrey M. Lange}
\ShortHeadings{Stationary Theory for Plateau Search}{Dukhovny and Lange}
\firstpageno{1}
\begin{document}
\title{A Stationary-Distribution Theory for Triplet-Based Plateau Search in Random Forest Ensemble-Size Selection}
\author{\name Andrey A. Dukhovny \email aadukhovny@sberbank.ru \\
        \addr Sberbank \\
        \addr Moscow 117997, Russia
        \AND
        \name Andrey M. Lange \email a.lange@skoltech.ru \\
        \addr Skolkovo Institute of Science and Technology (Skoltech), Moscow 121205, Russia, \\
        \addr Federal Research Center ``Computer Science and Control'' of Russian Academy of Sciences (FRC CSC RAS), Moscow 119333, Russia
}
\editor{To be assigned}
\maketitle
\begin{abstract}
The number of trees is a central computational parameter in Random Forests: increasing it reduces finite-ensemble variability but increases training and prediction cost.
Plateau-based tuning adapts this parameter through local comparisons of out-of-bag scores at a geometric triplet of tree counts.
After the remaining hyperparameters have stabilized, however, the central triplet point need not converge to a deterministic value; instead, it fluctuates around a stationary regime.

This paper develops a stationary-distribution theory for this process.
The central ensemble size \(B_t\) is modeled as a birth--death Markov chain on a geometric grid, and its stationary distribution is derived through local balance.
Under a leading centered folded-normal approximation, equilibrium equations are obtained for the original update rule and a symmetric modified variant, implying that the stationary center \(B_*=O(\varepsilon^{-2})\) as \(\varepsilon\downarrow0\).

The stationary spread is also characterized.
A local Gaussian approximation and a Fokker--Planck interpretation give grid-level variance constants.
After conversion to the ensemble-size scale, \(\sigma_{B,*}=O(\varepsilon^{-2})\), while the variance is \(O(\varepsilon^{-4})\).
The leading relative spread is independent of \(\varepsilon\) and controlled by the scale factor and update rule.
These results interpret plateau-based Random Forest tuning as a stochastic process rather than a deterministic stopping rule.
\end{abstract}
\section{Introduction}
\label{sec:introduction}
The setting is supervised learning on tabular data with \(n\) observations and \(p\) input features.
After standard preprocessing of categorical variables, such data can be represented by a design matrix \(X\in\mathbb{R}^{n\times p}\), whose rows correspond to observations and columns to features, together with a target vector \(y\).
Tree-based ensembles remain state-of-the-art for tabular data, consistently outperforming deep learning in many benchmark studies \citep{Grinsztajn2022Why, Shwartz2022Tabular,Borisov2024Deep}.
Among them, Gradient Boosting \citep{Friedman2001Greedy,Chen2016Xgboost} often delivers superior predictive accuracy, while Random Forest \citep{Breiman2001Random,Biau2016Random} offers greater stability.
This stability stems from averaging over many randomized trees: the Monte Carlo component of the ensemble prediction decreases as the number of trees grows.
The independence of trees also enables parallel training and the use of out-of-bag (OOB) scores, which provide an internal performance estimate without cross-validation.

Beyond prediction, Random Forests provide variable importance measures (VIMs), such as the Mean Decrease in Impurity (MDI), which naturally arise from the tree-building procedure \citep{Breiman2001Random,Louppe2013Understanding}.
These VIMs are widely used for feature selection, network inference, and scientific discovery \citep{Strobl2007Bias,Kursa2010Boruta,Ewald2024Guide}.
However, stabilizing VIMs in high-dimensional settings with correlated features may require substantially more trees than stabilizing the predictive score itself \citep{Lange2025optRF,Tolosi2011Classification}.
A well-calibrated predictive score is therefore a necessary, though not sufficient, condition for trustworthy VIMs.
This observation reinforces the motivation of the present line of work: before one can reliably assess variable importance, one must first obtain a Random Forest with stable and sufficiently accurate predictive performance.

The number of trees \(T\) is therefore a central computational parameter of Random Forests.
Increasing \(T\) reduces finite-ensemble variability, but it also increases training and prediction cost.
Standard hyperparameter optimization (HPO) methods, such as TPE \citep{Bergstra2011Algorithms} or Hyperband \citep{Li2018Hyperband}, require the user to specify a search range \([T_{\min},T_{\max}]\).
Because adding trees does not induce the usual overfitting behavior of many other hyperparameters, the selected value of \(T\) tends to be driven toward the upper boundary \(T_{\max}\).
Raising \(T_{\max}\) shifts the selected value further toward the boundary, offering no guarantee that the chosen bound is either sufficient or computationally efficient.
Early-stopping heuristics avoid an explicit upper bound by monitoring incremental score improvements, but they may stop prematurely when OOB score fluctuations make the observed improvement appear small.

In a recent paper, \citet{Porvatov2026How} introduced a triplet-based plateau search procedure that adapts the number of trees without requiring a fixed \(T_{\max}\).
At each HPO trial, the non-\(T\) hyperparameters are sampled in the usual way, while the ensemble size is represented by a geometric triplet \(L=B/\mathrm{sf}\), \(B\), and \(R=B\cdot\mathrm{sf}\), with a fixed scale factor \(\mathrm{sf}>1\).
The geometric construction keeps the relative separation between neighboring ensemble sizes approximately constant, since \(R-B=(\mathrm{sf}-1)B\) and hence \((R-B)/B=\mathrm{sf}-1\).
This is important because, for a fixed additive step \(R=B+\Delta\), the relative separation satisfies \((R-B)/B=\Delta/B\to0\) as \(B\to\infty\), making neighboring forests increasingly difficult to distinguish by their OOB scores.
Fixed increments of this type, for example \(\Delta=10\), were used in earlier ensemble-size studies \citep{Latinne2001Limiting,Lange2025optRF}.

Thus, the scale factor \(\mathrm{sf}\) acts as a resolution parameter for the plateau comparison.
The forest is trained sequentially up to \(R\) trees, and the OOB scores at \(L\), \(B\), and \(R\) are recorded.
The relative score gaps
\begin{equation}
\label{eq:rel_diff_intro}
d_L
=
\frac{|S_B-S_L|}{|S_B|},
\qquad
d_R
=
\frac{|S_R-S_B|}{|S_B|}
\end{equation}
indicate how close the central ensemble size \(B\) is to the plateau region.
They are compared with a user-specified tolerance \(\varepsilon\), typically of the order \(10^{-3}\).

If both inequalities \(d_L\le\varepsilon\) and \(d_R\le\varepsilon\) hold, the ensemble is considered stable but potentially unnecessarily large, and the triplet is shifted left for the next trial.
In the original rule, if \(d_R>\varepsilon\), the improvement from \(B\) to \(R\) is still above the tolerance, and the current ensemble is treated as insufficient; the triplet is therefore shifted right.
In the remaining original-rule case, \(d_L>\varepsilon\) and \(d_R\le\varepsilon\), the triplet remains at the current level.
A symmetric modified rule is also analyzed below; in this variant, the mixed case \(d_L\le\varepsilon<d_R\) is assigned to staying rather than to a right shift.
Importantly, trials that trigger a right shift are not considered later for selecting the trial with the highest score, because they are deemed too unstable to be trusted.
This adaptive mechanism eliminates the need for an arbitrary upper bound \(T_{\max}\) and avoids the systematic underestimation bias of one-shot early stopping.
Moreover, it jointly optimizes the ensemble size with other hyperparameters such as tree depth and \(m_{\mathrm{try}}\), recognizing that the required number of trees interacts with the remaining hyperparameters \citep{Probst2019Tunability, Bernard2009Influence}.

However, this adaptivity changes the nature of the problem.
The central triplet point \(B_t\) evolves across HPO trials according to random OOB score comparisons.
This randomness has several sources.
First, the OOB scores are computed from finite data and therefore inherit sampling variability.
Second, even for fixed hyperparameters and fixed data, Random Forest training involves algorithmic randomness from bootstrap sampling and feature subsampling, which produces finite-ensemble fluctuations of the OOB score.
In addition, the non-\(T\) hyperparameters are sampled by the HPO procedure, so changes in \(B_t\) are coupled with the stochastic exploration of the remaining search space.
\begin{figure}[t]
\centering
\includegraphics[width=\textwidth]{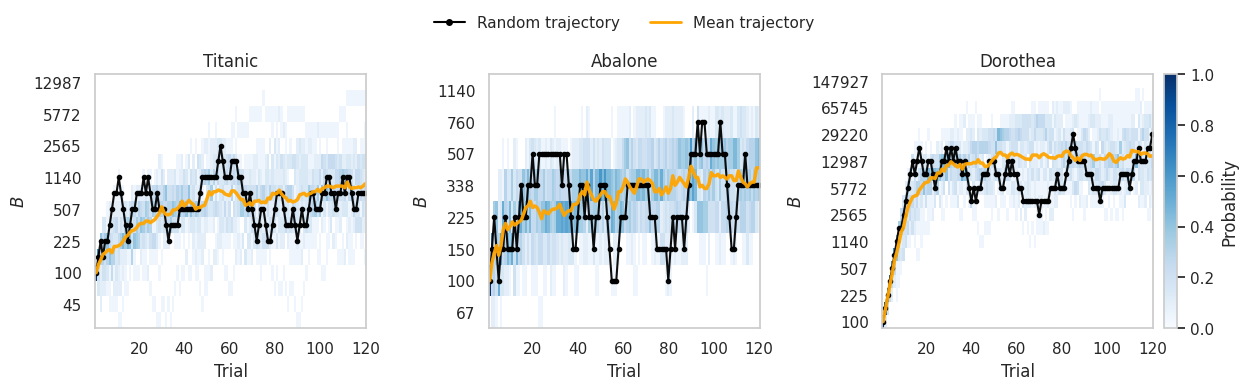}
\caption{
Examples of random and mean trajectories of the central triplet point \(B_t\) across HPO trials for three data sets, reproduced from \protect\citet{Porvatov2026How}.
The mean trajectory is computed over 20 runs, and the background colormap shows the empirical frequency of the corresponding tree count at each trial, illustrating the emergence of a stationary regime.
Here \(T_0=100\) and \(\mathrm{sf}=1.5\).
In the left panel, \(\varepsilon=3\cdot10^{-3}\), while in the other two panels, \(\varepsilon=10^{-3}\).
}
\label{fig:trajectories}
\end{figure}

The triplet-based plateau algorithm partially regularizes this randomness.
Unlike one-shot early stopping, it updates the central ensemble size across trials, so left and right corrections can compensate over time.
Nevertheless, the process remains random.
Consequently, the selected number of trees exhibits substantial run-to-run variability across independent optimization runs with different random seeds.
This variability limits the reliability of a single-point estimate of \(T\): a practitioner who runs the same HPO procedure twice may obtain noticeably different recommendations for the ensemble size.
This variability is not merely a nuisance, but a fundamental property of the adaptive tuning process.
The selected number of trees should therefore be viewed not as a deterministic quantity, but as a random variable induced by the optimization dynamics.

Figure~\ref{fig:trajectories} illustrates this behavior.
After an initial transient phase, the central triplet point \(B_t\) does not converge to a fixed deterministic value, but fluctuates within a problem-dependent range that is naturally interpreted as a stationary regime.
The visibly non-negligible amplitude of these fluctuations motivates the variance analysis below, which shows that the stationary standard deviation on the ensemble-size scale can be a sizeable fraction of the stationary mean.
The tolerance \(\varepsilon\) controls how small the relative OOB score gaps in \eqref{eq:rel_diff_intro} must be before the plateau tests are regarded as passed.
The equilibrium center of this regime is denoted by \(B_*\).
As \(\varepsilon\) decreases, the plateau condition becomes more stringent, so the equilibrium level shifts toward larger ensemble sizes.
This shift also increases the absolute size of the fluctuations when measured in the number of trees.
At the same time, the OOB score-difference noise in \eqref{eq:rel_diff_intro} becomes smaller at larger ensemble sizes because the finite-ensemble variance decays with the number of trees.

This paper develops a theoretical description of this stationary regime.
The analysis assumes that the non-\(T\) hyperparameters have effectively stabilized and isolates the stochastic evolution of the central ensemble size \(B_t\).
Under this reduced description, the plateau procedure induces a birth--death Markov chain on the geometric grid \(B_j=T_0\cdot\mathrm{sf}^j\).
The goal is not to construct a practical estimator from a finite HPO trajectory, but to characterize the population-level stationary quantities of this process: its stationary distribution, equilibrium center, and local spread.
This separates the theoretical question studied here from the separate practical problem of estimating these quantities from finite trajectories.

The analysis proceeds in three steps.
First, the transition probabilities induced by the original plateau update rule and by a symmetric modified variant are derived.
These probabilities are expressed through plateau-pass probabilities for the left and right OOB score gaps.
Second, the finite-ensemble variance scaling of the signed relative gaps is used to approximate these pass probabilities by a leading centered folded-normal model.
This yields explicit equilibrium equations and shows that the stationary center satisfies \(B_*=O(\varepsilon^{-2})\) as \(\varepsilon\downarrow0\).
Third, the local shape of the stationary distribution is analyzed.
A product-form local-balance argument gives a local Gaussian approximation, and an equivalent Fokker--Planck interpretation yields explicit variance constants for both update rules.
After converting from the grid-level scale to the ensemble-size scale, the stationary standard deviation also satisfies \(\sigma_{B,*}=O(\varepsilon^{-2})\), while the corresponding variance is \(O(\varepsilon^{-4})\).

The main contributions of this paper are as follows.
First, triplet-based plateau search is formulated as a birth--death Markov chain on a geometric ensemble-size grid.
Second, the product-form stationary distribution is derived and conditions under which it is normalizable are established.
Third, under a leading folded-normal approximation, explicit equilibrium equations are obtained for the original and modified update rules, yielding the scaling law \(B_*=O(\varepsilon^{-2})\).
Fourth, local stationary variance constants are derived, interpreted through a drift--diffusion representation, and used to show that the ensemble-size standard deviation satisfies \(\sigma_{B,*}=O(\varepsilon^{-2})\).
This appears to be the first work to explicitly model the Random Forest ensemble size selected by an adaptive HPO procedure as a stochastic process and to characterize its stationary behavior.
Together, these results provide a theoretical basis for interpreting plateau-based Random Forest tuning as a stationary stochastic process rather than as a deterministic stopping rule.

The present paper is an analytical study. 
It characterizes the population-level stationary quantities of the plateau process and does not attempt to construct a finite-trajectory estimator. 
Empirical calibration, synthetic Markov-chain simulations, and moment-based estimators from finite HPO trajectories are natural follow-up problems and are discussed in Section~\ref{sec:discussion}.
The remainder of the paper is organized as follows.
Section~\ref{sec:related_work} reviews related work on selecting the number of trees, plateau and stopping rules, and noisy hyperparameter optimization.
Section~\ref{sec:plateau_process} reformulates triplet-based plateau search as a stochastic update process and derives the signed-gap variance asymptotics used later.
Section~\ref{sec:markov} develops the birth--death Markov model, characterizes its stationary distribution, and derives the equilibrium and variance scaling laws.
Section~\ref{sec:discussion} discusses the implications and limitations of the theory, including the role of the geometric grid and directions for empirical follow-up.
Appendix~\ref{app:proofs} contains the proofs.
\section{Related Work}
\label{sec:related_work}
\subsection{Selecting the Number of Trees in Random Forest}
The challenge of selecting the number of trees \(T\) has been studied extensively.
Early empirical work by \citet{Oshiro2012How} and \citet{Genuer2008Random} confirmed that the predictive score plateaus as \(T\) increases, with \citet{Oshiro2012How} relating the required \(T\) to data set density \(\log_p n\).
\citet{Cuzzocrea2013Information} and \citet{Demidova2020Approach} proposed heuristic mappings for \(T\) based on information-theoretic measures and data set size, respectively.
Statistical stopping criteria were developed by \citet{Latinne2001Limiting}, who used a McNemar test, and by \citet{Hernandez2013How}, who derived an asymptotic estimate for the ensemble size needed to match an infinite forest.
\citet{Lopes2019Estimating} analyzed the algorithmic variance of randomized ensembles, showing that the conditional error variance decays as \(O(T^{-1})\) and discussing relative stopping conditions.
More recently, \citet{Lange2025optRF} modeled the relationship between \(T\) and the stability of variable importance measures using a logistic function, extrapolating to very large ensemble sizes.

These works provide valuable guidelines for determining a sufficient number of trees.
Most of them, however, treat \(T\) as a deterministic target for a fixed learning problem or for a fixed set of hyperparameters.
In contrast, the plateau search procedure studied here adapts \(T\) across HPO trials, and the central ensemble size evolves as a stochastic process driven by OOB score comparisons.
The focus here is therefore not a single stopping time or a single sufficient value of \(T\), but the stationary distribution induced by the adaptive triplet update.
\subsection{Using the HPO Trace}
A growing body of research argues that discarding all but the single best HPO trial is wasteful.
\citet{Hutter2014Efficient} introduced fANOVA, a post-hoc method that uses the entire history of HPO runs to decompose the variance of the observed performance, identifying which hyperparameters are important and how they interact.
\citet{Caruana2004Ensemble} demonstrated that constructing an ensemble from a library of models generated during a single HPO run can outperform selecting the single best model.
Both approaches exploit information contained in the optimization trace rather than relying only on the best observed configuration.

The goal of the present work is different.
The present work does not aggregate several fitted models, nor does it estimate global hyperparameter importances.
Instead, the trajectory of a single hyperparameter, the number of trees, is used as the object of analysis.
The central point \(B_t\) of the plateau triplet is treated as a stochastic process, and the HPO trace motivates studying its stationary center and spread.
This view is especially relevant for Random Forests because the effect of \(T\) is primarily computational and variance-reducing rather than a conventional overfitting tradeoff.
\subsection{Noisy and Robust Hyperparameter Optimization}
Noisy observations are a central concern in Bayesian optimization (BO).
When function evaluations are noisy, standard BO algorithms can be misled by random fluctuations.
\citet{Letham2019Constrained} proposed a modification of the expected improvement acquisition function that integrates over the posterior distribution of noisy observations.
Such methods are designed to improve decision making in the presence of observation noise, often by modifying the acquisition function or the identification step.

In the Random Forest plateau setting, the noise has a specific structure.
As shown in \citet{Porvatov2026How}, the variance of relative OOB score differences \eqref{eq:rel_diff_intro} decays as \(O(1/T)\), or equivalently their standard deviation scales as \(O(T^{-1/2})\).
Thus, the uncertainty is heteroscedastic and directly tied to the ensemble size.
A generic Gaussian process (GP) surrogate over \(T\) would need to account for this structure, while also separating true saturation of the forest from finite-ensemble OOB fluctuations.
More importantly, the objective of noisy BO methods is typically to identify a single best configuration under uncertainty.
The objective here is instead to characterize the distribution of the ensemble size produced by an adaptive plateau process.

Related distinctions arise in robust and risk-seeking Bayesian optimization.
\citet{Iwazaki2024Risk} formulated a risk-seeking BO problem, aiming to find the single best possible reward under uncontrollable environmental noise.
For a monotone Random Forest score, such a risk-seeking strategy would tend to favor larger values of \(T\) and does not by itself define a sufficient ensemble size.
\citet{BelandNair2017} considered robust optimization under uncertainty by building a GP surrogate for an integral of the objective over uncontrollable variables.
These methods address robustness or risk preferences in optimization, whereas the present paper studies the stationary stochastic dynamics generated by a specific adaptive tuning rule.
\subsection{Position of the Present Work}
The present paper is closest in spirit to work on stopping rules and sufficient ensemble size, but it changes the object of analysis.
Rather than asking when a Random Forest should stop growing, the question considered here is where the balance between upward and downward ensemble-size moves is established on the geometric grid, and how broadly the process fluctuates around this balance point.
This perspective is motivated by the empirical observation that the central triplet point need not settle at a single deterministic value.
It may instead fluctuate around a problem-dependent region after an initial transient phase.

The contribution is theoretical.
The plateau update is formulated as a birth--death Markov chain, its product-form stationary distribution is derived, and conditions for existence and uniqueness of the stationary distribution are established.
Under a leading folded-normal approximation for the OOB score-gap probabilities, explicit balance equations are obtained for the original update rule and for a symmetric modified variant.
These equations yield the scaling law \(B_*=O(\varepsilon^{-2})\) for the stationary center.
Local variance constants are further derived, showing that the stationary standard deviation on the ensemble-size scale also satisfies \(\sigma_{B,*}=O(\varepsilon^{-2})\).

Thus, for a user-specified tolerance such as \(\varepsilon=10^{-3}\), decreasing the tolerance increases both the stationary mean ensemble size and the absolute magnitude of its fluctuations.
At the same time, their leading ratio remains controlled by the scale factor and the update rule, rather than by \(\varepsilon\).
For commonly used scale factors, this relative spread can be substantial: for example, it is approximately one half of the stationary mean for \(\mathrm{sf}=1.5\).
This shows that the variability of the selected number of trees is not only a finite-sample nuisance, but a structural feature of the stationary plateau process.

This distinguishes the paper from post-hoc heuristics, robust BO methods, and model-aggregation approaches.
Those methods aim to improve selection, prediction, or robustness in an HPO workflow.
Here, the aim is to understand the stochastic process induced by the plateau rule itself.
The derived relation between the stationary center and the stationary spread can also serve as a theoretical basis for future trajectory-based estimators of the mean number of trees, including moment-based approaches such as the generalized method of moments.
The construction and empirical benchmarking of such estimators are left for separate work.
\section{Plateau Search as a Stochastic Process}
\label{sec:plateau_process}
The triplet-based plateau search was introduced in \citet{Porvatov2026How} as an adaptive mechanism for selecting the number of trees without specifying an explicit upper bound \(T_{\max}\).
The full algorithmic description was summarized in the Introduction.
This section extracts the part of the construction that is needed for the stochastic analysis: after the non-\(T\) hyperparameters have effectively stabilized, the central triplet point \(B_t\) evolves across HPO trials according to random threshold comparisons of OOB scores.
\subsection{Stochastic Plateau Updates}
\label{subsec:stochastic_updates}
The full plateau-based HPO procedure jointly samples the non-\(T\) hyperparameters by TPE and updates the ensemble size through a triplet rule.
Since the present analysis focuses on the stationary regime after the non-\(T\) hyperparameters have effectively stabilized, only the ensemble-size update component is isolated.
At trial \(t\), given the current central point \(B_t\), the triplet points are
\(L_t=B_t/\mathrm{sf}\), \(B_t\), and \(R_t=B_t\cdot\mathrm{sf}\).
For the sampled non-\(T\) hyperparameter configuration, a nested forest is trained up to \(R_t\) trees, and the OOB scores \(S_{L_t}\), \(S_{B_t}\), and \(S_{R_t}\) are recorded.

The plateau rule is based on the two absolute relative gaps
\[
d_{L,t}
=
\left|
\frac{S_{B_t}-S_{L_t}}{S_{B_t}}
\right|,
\qquad
d_{R,t}
=
\left|
\frac{S_{R_t}-S_{B_t}}{S_{B_t}}
\right|.
\]
The next central point is then updated according to the four threshold cases
\[
B_{t+1}
=
\begin{cases}
B_t\cdot\mathrm{sf},
&
d_{L,t}>\varepsilon,\quad d_{R,t}>\varepsilon,
\\[1mm]
B_t,
&
d_{L,t}>\varepsilon,\quad d_{R,t}\le\varepsilon,
\\[1mm]
B_t/\mathrm{sf},
&
d_{L,t}\le\varepsilon,\quad d_{R,t}\le\varepsilon,
\\[1mm]
B_t\cdot\mathrm{sf}\ \text{in the original rule, or } B_t\ \text{in the modified rule},
&
d_{L,t}\le\varepsilon,\quad d_{R,t}>\varepsilon .
\end{cases}
\]
Thus, the original algorithm and the modified variant considered below differ only in the mixed case where the left plateau test passes while the right plateau test fails.
The modified rule assigns this case to a stay decision, which will lead to a more symmetric transition structure in the Markov model.

For the analysis below, it is useful to view the absolute gaps in \eqref{eq:rel_diff_intro} as absolute values of their signed counterparts,
\begin{equation}
\label{eq:signed_relative_gaps}
\frac{S_{B_t}-S_{L_t}}{S_{B_t}},
\qquad
\frac{S_{R_t}-S_{B_t}}{S_{B_t}} .
\end{equation}
The first quantity measures the local score change from the left neighbor \(L_t\) to the central point \(B_t\), while the second measures the change from \(B_t\) to the right neighbor \(R_t\).

Because the OOB scores are random, the quantities \(d_{L,t}\) and \(d_{R,t}\) are random even when the data and the non-\(T\) hyperparameters are fixed.
Consequently, the update of \(B_t\) is stochastic.
The unambiguous cases correspond to shifting right, staying, or shifting left according to the plateau logic.
The remaining mixed case can be assigned either to a right shift or to a stay decision.
Both variants are handled in the Markov formulation below; the distinction affects only the transition probabilities, not the asymptotic score model used to derive them.
\subsection{Signed-Gap Variance Asymptotics}
\label{subsec:signed_gap_asymptotics}
The finite-ensemble asymptotic model developed in \citet{Porvatov2026How} is used.
Conditionally on the training data \(D\), let \(\mu_T=\mathbb{E}[S_T\mid D]\) denote the expected OOB score of a Random Forest with \(T\) trees.
The conditional mean score is assumed to converge to an infinite-forest limit according to
\begin{equation}
\label{eq:score_convergence}
\mu_T
=
S_\infty + cT^{-\gamma} + o(T^{-\gamma}),
\qquad T\to\infty,
\end{equation}
where \(S_\infty\ne 0\), \(c\ne 0\), and \(\gamma>0\).
The stronger condition \(\gamma>1/2\) will be used below when deriving the centered folded-normal approximation for plateau-pass probabilities.

The variance calculation follows the same finite-ensemble Gaussian approximation used in \citet{Porvatov2026How}.
For the variance asymptotics themselves, however, only its second-order consequences are needed: the \(O(T^{-1})\) decay of algorithmic variance and the covariance scaling for nested warm-start forests, formalized in \eqref{eq:finite_ensemble_covariance_scaling}.
The variance component in \eqref{eq:finite_ensemble_covariance_scaling} is motivated by finite-ensemble variance results for randomized ensembles, in particular the \(O(T^{-1})\) decay analyzed by \citet{Lopes2019Estimating}.
The covariance component in \eqref{eq:finite_ensemble_covariance_scaling} is an additional nested-forest approximation tailored to the warm-start construction used by the plateau algorithm: when \(T_1<T_2\), the larger forest contains the smaller forest as a prefix, leading to the leading-order covariance \(v/T_2\).
\begin{restatable}[Signed relative gap variance asymptotics]{myproposition}{signedgapvarianceasymptotics}
\label{prop:signed_gap_variance_asymptotics}
Let \(L=B/\mathrm{sf}\) and \(R=B\cdot\mathrm{sf}\), with \(\mathrm{sf}>1\).
Assume \eqref{eq:score_convergence} and the finite-ensemble covariance scaling
\begin{equation}
\label{eq:finite_ensemble_covariance_scaling}
\operatorname{Var}[S_T\mid D]\sim \frac{v}{T},
\qquad
\operatorname{Cov}[S_{T_1},S_{T_2}\mid D]\sim \frac{v}{T_2},
\qquad T_1<T_2,
\end{equation}
for some problem-dependent constant \(v>0\).
Then, as \(B\to\infty\),
\begin{align}
\label{eq:left_gap_variance}
\operatorname{Var}\left[
\frac{S_B-S_L}{S_B}
\;\middle|\;D
\right]
&\sim
\frac{v}{S_\infty^2}
\frac{\mathrm{sf}-1}{B},\\
\label{eq:right_gap_variance}
\operatorname{Var}\left[
\frac{S_R-S_B}{S_B}
\;\middle|\;D
\right]
&\sim
\frac{v}{S_\infty^2}
\frac{1-\mathrm{sf}^{-1}}{B}.
\end{align}
\end{restatable}

The resulting right-gap variance in \eqref{eq:right_gap_variance} was derived in \citet{Porvatov2026How}.
The corresponding left-gap variance in \eqref{eq:left_gap_variance} is derived here by the same nested-forest covariance argument; the proof is given in Appendix~\ref{app:proofs}.
Both expressions are needed because the Markov model below uses both the left and right plateau tests.
The factor difference between \eqref{eq:left_gap_variance} and \eqref{eq:right_gap_variance} reflects only the fact that the left comparison involves the smaller ensemble \(L=B/\mathrm{sf}\), whose finite-forest variance is larger.
It does not assume any scaling relation between the scores \(S_L\), \(S_B\), and \(S_R\) themselves.
In particular, \((\mathrm{sf}-1)/(1-\mathrm{sf}^{-1})=\mathrm{sf}\), so the left signed relative gap has asymptotically \(\mathrm{sf}\) times larger variance than the right signed relative gap under the nested covariance approximation.
\subsection{Absolute Gaps and Plateau-Pass Probabilities}
\label{subsec:absolute_plateau_probs}
The actual plateau rule uses the absolute values of the signed relative gaps in \eqref{eq:signed_relative_gaps}.
For a generic central value \(B\), define the left and right plateau-pass probabilities by
\begin{equation}
\label{eq:plateau_pass_probabilities_B}
\alpha_L(B;\varepsilon) = 
\mathbb{P}\left[
\left|
\frac{S_B-S_{B/\mathrm{sf}}}{S_B}
\right|
\le \varepsilon
\;\middle|\;D
\right], \qquad 
\alpha_R(B;\varepsilon) = 
\mathbb{P}\left[
\left|
\frac{S_{B\cdot\mathrm{sf}}-S_B}{S_B}
\right|
\le \varepsilon
\;\middle|\;D
\right].
\end{equation}

\begin{figure}[t]
\centering
\includegraphics[width=\linewidth]{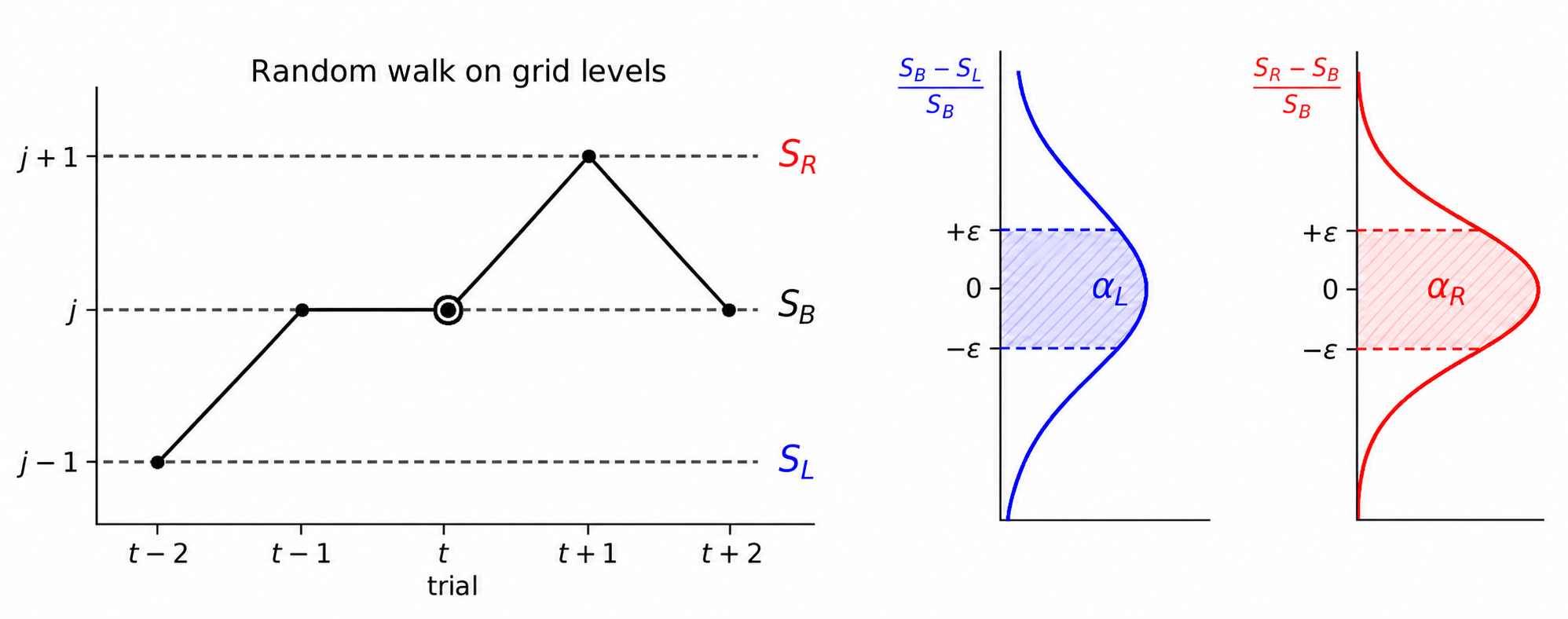}
\caption{Triplet update and schematic signed-gap distributions; shaded regions represent \(\alpha_L\) and \(\alpha_R\).}
\label{fig:triplet_signed_gap_schematic}
\end{figure}

Figure~\ref{fig:triplet_signed_gap_schematic} gives a schematic interpretation of
\(\alpha_L(B;\varepsilon)\) and \(\alpha_R(B;\varepsilon)\).
The shaded regions correspond to the events that the corresponding signed relative gaps fall inside the tolerance interval \([-\varepsilon,\varepsilon]\).
This notation is intentionally distribution-agnostic: the probabilities in \eqref{eq:plateau_pass_probabilities_B} may be evaluated using a folded-normal approximation, an empirical or bootstrap approximation, or another model for the OOB score fluctuations.

To obtain explicit analytical formulas, a stronger approximation than the variance calculation above is now imposed.
Proposition~\ref{prop:signed_gap_variance_asymptotics} uses only the second-order asymptotics of the signed gaps in \eqref{eq:signed_relative_gaps}, whereas the next result assumes an approximate Gaussian law for these transformed signed relative gaps themselves.
\begin{restatable}[Plateau-pass probability asymptotics]{myproposition}{plateaupassprobabilityasymptotics}
\label{prop:plateau_pass_probability_asymptotics}
Let \(L=B/\mathrm{sf}\) and \(R=B\cdot\mathrm{sf}\), and let \(s_L^2(B)\) and \(s_R^2(B)\) denote the leading variance scales in
\eqref{eq:left_gap_variance} and \eqref{eq:right_gap_variance}, respectively.
Assume that the signed relative gaps admit the conditional Gaussian approximations
\[
\left.\frac{S_B-S_L}{S_B}\;\right|\; D
\approx
\mathcal{N}\bigl(m_L(B),s_L^2(B)\bigr),
\qquad
\left.\frac{S_R-S_B}{S_B}\;\right|\; D
\approx
\mathcal{N}\bigl(m_R(B),s_R^2(B)\bigr).
\]
Assume the tail model \eqref{eq:score_convergence} with \(\gamma>1/2\).
Then the conditional means of the signed relative gaps satisfy
\(m_L(B)=O(B^{-\beta})\) and \(m_R(B)=O(B^{-\beta})\), where
\(\beta=\min\{\gamma,1\}>1/2\), and, to leading order,
\begin{align}
\label{eq:alpha_left_asymptotic}
\alpha_L(B;\varepsilon)
&=
2\Phi\left(\frac{\varepsilon}{s_L(B)}\right)-1
+
O\left(B^{1-2\beta}\right),\\
\label{eq:alpha_right_asymptotic}
\alpha_R(B;\varepsilon)
&=
2\Phi\left(\frac{\varepsilon}{s_R(B)}\right)-1
+
O\left(B^{1-2\beta}\right),
\end{align}
where \(\Phi(\cdot)\) is the standard normal cumulative distribution function.
\end{restatable}

The proof is given in Appendix~\ref{app:proofs}.
Substituting the variance scales from \eqref{eq:left_gap_variance}--\eqref{eq:right_gap_variance} into \eqref{eq:alpha_left_asymptotic}--\eqref{eq:alpha_right_asymptotic} gives the leading centered folded-normal approximations
\begin{equation}
\label{eq:alpha_leading}
\alpha_L(B;\varepsilon)
\approx
2\Phi\left(
\varepsilon
\sqrt{
\frac{S_\infty^2 B}{v(\mathrm{sf}-1)}
}
\right)-1,\qquad
\alpha_R(B;\varepsilon)
\approx
2\Phi\left(
\varepsilon
\sqrt{
\frac{S_\infty^2 B}{v(1-\mathrm{sf}^{-1})}
}
\right)-1 .
\end{equation}

For a signed gap \(X_B\) with conditional mean \(m(B)\) and standard deviation \(s(B)\), the non-centered folded-normal approximation is
\begin{equation}
\label{eq:folded_normal_noncentered}
\mathbb{P}\left[
|X_B|\le\varepsilon
\;\middle|\;D
\right]
\approx
\Phi\left(\frac{\varepsilon-m(B)}{s(B)}\right)
-
\Phi\left(\frac{-\varepsilon-m(B)}{s(B)}\right).
\end{equation}
The asymptotics in \eqref{eq:alpha_left_asymptotic}--\eqref{eq:alpha_right_asymptotic} follow by expanding \eqref{eq:folded_normal_noncentered} in powers of the normalized mean \(m(B)/s(B)\).
The first-order terms cancel because the two Gaussian CDF terms enter with opposite signs, so the first nonzero correction is quadratic,
that is, of order $O(m^2(B)/s^2(B))$.
The behavior \(m(B)=O(B^{-\beta})\), where \(\beta=\min\{\gamma,1\}>1/2\), follows from a second-order delta-method expansion.
Together with \(s(B)\asymp B^{-1/2}\), this implies that \(m(B)/s(B)\to0\), and hence the correction due to the nonzero conditional mean vanishes asymptotically.
Thus, the approximations in \eqref{eq:alpha_leading} are centered leading-order folded-normal approximations.

Note that the leading terms in \eqref{eq:alpha_leading} do not involve \(\gamma\); they are determined only by the variance scales in \eqref{eq:left_gap_variance}--\eqref{eq:right_gap_variance}.
The condition \(\gamma>1/2\) ensures the vanishing of \(m_L(B)/s_L(B)\) and \(m_R(B)/s_R(B)\).
Moreover, the approximation does not require any specific limiting behavior of the ratios \(\varepsilon/s_L(B)\) and \(\varepsilon/s_R(B)\).
The tolerance \(\varepsilon\) is a user-controlled parameter, and decreasing it shifts the stationary regime toward larger ensembles, where the noise scales \(s_L(B)\) and \(s_R(B)\) are smaller.
However, the correction due to the nonzero conditional means is bounded by \(O(B^{1-2\beta})\) uniformly in \(\varepsilon/s(B)\).

The asymptotic forms in \eqref{eq:alpha_left_asymptotic}--\eqref{eq:alpha_right_asymptotic}, and their explicit leading versions in \eqref{eq:alpha_leading}, provide the probabilistic input for the Markov model.
As \(B\) increases, the standard deviations \(s_L(B)\) and \(s_R(B)\) decrease as \(B^{-1/2}\), so the leading plateau-pass probabilities in \eqref{eq:alpha_leading} increase toward one.
This creates an inward drift: small ensembles tend to shift right because the plateau tests often fail, whereas sufficiently large ensembles tend to shift left because both tests are likely to pass.
In the next section, this intuition is formalized as a birth--death Markov chain and its stationary distribution is analyzed.
\section{Markov Model and Stationary Distribution}
\label{sec:markov}
The stochastic plateau updates are now converted into a birth--death Markov chain.
The central point \(B_t\) is assumed to evolve on the geometric grid
\(T_j=T_0 \cdot \mathrm{sf}^j\), \(j=0,1,\ldots\),
where \(T_0>0\) is the initial ensemble size.
Thus, for an interior level, the event \(B_t=T_j\) means that at trial \(t\) the central triplet point is located at level \(j\) of the fixed geometric grid, with
\(L_t=T_{j-1}=B_t/\mathrm{sf}\) and \(R_t=T_{j+1}=B_t\cdot\mathrm{sf}\), up to rounding to integer tree counts.
When no ambiguity is possible, the notation \(B_j=T_j\) is also used for the central ensemble size associated with grid level \(j\).
Although the implementation may occasionally move to negative grid indices \(j=-1,-2,\ldots\), this can only occur while the rounded tree counts remain positive and distinct; such lower-boundary effects are rare and do not affect the large-\(B\) stationary analysis below.

On this grid, for simplicity, write
\(\alpha_{L,j}=\alpha_L(T_j;\varepsilon)\) and
\(\alpha_{R,j}=\alpha_R(T_j;\varepsilon)\),
where \(\alpha_L\) and \(\alpha_R\) were defined in
\eqref{eq:plateau_pass_probabilities_B}.
Thus, \(\alpha_{L,j}\) is the probability that the left plateau test passes when the central point is \(B_t=T_j\), and \(\alpha_{R,j}\) is the corresponding probability for the right plateau test.
When there is no ambiguity, the argument \(\varepsilon\) is suppressed.
\subsection{Plateau Cases and Update Variants}
\label{subsec:plateau_cases}
The update rule is determined by the outcomes of the left and right plateau tests.
Here ``pass'' means that the corresponding absolute relative gap is at most \(\varepsilon\), and ``fail'' means that it exceeds \(\varepsilon\).
The four possible cases are shown in Table~\ref{tab:plateau_cases}.
\begin{table}[t]
\centering
\begin{tabular}{llll}
\toprule
Left gap & Right gap & Original update & Modified update \\
\midrule
\(\left|(S_B-S_L)/S_B\right|>\varepsilon\) 
& \(\left|(S_R-S_B)/S_B\right|>\varepsilon\) 
& shift right 
& shift right \\

\(\left|(S_B-S_L)/S_B\right|>\varepsilon\) 
& \(\left|(S_R-S_B)/S_B\right|\le\varepsilon\) 
& stay 
& stay \\
\(\left|(S_B-S_L)/S_B\right|\le\varepsilon\) 
& \(\left|(S_R-S_B)/S_B\right|>\varepsilon\) 
& shift right 
& stay \\

\(\left|(S_B-S_L)/S_B\right|\le\varepsilon\) 
& \(\left|(S_R-S_B)/S_B\right|\le\varepsilon\) 
& shift left 
& shift left \\
\bottomrule
\end{tabular}
\caption{Plateau-test outcomes and update decisions.}
\label{tab:plateau_cases}
\end{table}
The only difference between the two variants is the mixed case in which the left test passes but the right test fails.
In the original algorithm of \citet{Porvatov2026How}, this case is assigned to a right shift.
In the modified variant, it is assigned to staying at the current level.
The modified rule is more symmetric: a left shift occurs only when both tests pass, a right shift occurs only when both tests fail, and mixed evidence leads to no move.
\subsection{Transition Probabilities}
\label{subsec:transition_probabilities}
The transition probabilities of the Markov chain are introduced as
\begin{equation*}
p_j^-=\mathbb{P}(B_{t+1}=T_{j-1}\mid B_t=T_j),
\qquad
p_j^+=\mathbb{P}(B_{t+1}=T_{j+1}\mid B_t=T_j),
\end{equation*}
and \(p_j^0=1-p_j^- - p_j^+\).
Here \(p_j^-\) is the probability of a left shift, \(p_j^+\) is the probability of a right shift, and \(p_j^0\) is the probability of staying at the same level.

The exact transition probabilities are determined by the joint distribution of the left and right plateau tests.
For explicit analytical formulas, the factorized approximation is used, in which the two tests are treated as conditionally independent at the same level.
Under this approximation, the original update rule gives
\begin{equation}
\label{eq:transition_original}
p_j^-
\approx
\alpha_{L,j}\alpha_{R,j},
\qquad
p_j^+
\approx
1-\alpha_{R,j},
\qquad
p_j^0
\approx
\alpha_{R,j}(1-\alpha_{L,j}).
\end{equation}
Indeed, a left shift occurs when both tests pass, while a right shift occurs whenever the right test fails.

For the modified update rule, the transition probabilities become
\begin{equation}
\label{eq:transition_modified}
p_j^-
\approx
\alpha_{L,j}\alpha_{R,j},
\qquad
p_j^+
\approx
(1-\alpha_{L,j})(1-\alpha_{R,j}),
\qquad
p_j^0
\approx
\alpha_{R,j}(1-\alpha_{L,j})
+
\alpha_{L,j}(1-\alpha_{R,j}).
\end{equation}
In this case, the chain moves right only when both tests fail, moves left only when both tests pass, and remains at the current level in the two mixed cases.
If the dependence between the two plateau tests is modeled explicitly, the products in
\eqref{eq:transition_original}--\eqref{eq:transition_modified}
should be replaced by the corresponding joint probabilities.
The birth--death analysis below remains unchanged after this replacement.

In the remainder of the paper, the auxiliary birth--death chains defined by the factorized transition probabilities in \eqref{eq:transition_original} and \eqref{eq:transition_modified} are analyzed.
Within these auxiliary chains, the displayed transition probabilities are treated as exact once the plateau-pass probabilities \(\alpha_{L,j}\) and \(\alpha_{R,j}\) are specified.
In contrast, the leading centered folded-normal relation \eqref{eq:alpha_leading} is an approximation linking these probabilities to the OOB score-gap model.
For the equilibrium analysis, this approximation is used to obtain leading asymptotic balance equations.
For the local variance analysis, the corresponding auxiliary chain is introduced by treating \eqref{eq:alpha_leading} as exact on the grid \(B_j=T_0\cdot\mathrm{sf}^j\), which allows the local derivatives of the transition probabilities to be evaluated explicitly.
\subsection{Stationary Distribution}
\label{subsec:stationary_distribution}
The transition probabilities in \eqref{eq:transition_original} or \eqref{eq:transition_modified} define a birth--death Markov chain on the levels \(j=0,1,\ldots\) of the geometric grid.
If the limiting distribution exists, it is denoted by
\(\pi_j=\lim_{t\to\infty}\mathbb{P}(B_t=T_j)\).
Any such limiting distribution must satisfy the stationary Kolmogorov--Chapman equations.
The existence and normalizability of this distribution are established in the next subsection.

The lower-boundary convention is \(p_0^-=0\).
The stationary Kolmogorov--Chapman equations are
\begin{align*}
\pi_0
&=
\pi_0p_0^0
+
\pi_1p_1^-,
\\
\pi_j
&=
\pi_{j-1}p_{j-1}^+
+
\pi_jp_j^0
+
\pi_{j+1}p_{j+1}^-,
\qquad j > 0.
\end{align*}
These equations follow from the law of total probability: to be at level \(j\) after one step, the chain must have shifted right from \(j-1\), stayed at \(j\), or shifted left from \(j+1\).
Since \(p_0^0=1-p_0^+\) and \(p_j^0=1-p_j^- -p_j^+\) for \(j\ge1\), these equations are equivalently written as
\begin{align*}
\pi_0p_0^+
&=
\pi_1p_1^-,
\\
\pi_j(p_j^-+p_j^+)
&=
\pi_{j-1}p_{j-1}^+
+
\pi_{j+1}p_{j+1}^-,
\qquad j > 0.
\end{align*}
Substituting the first equation into the second one for \(j=1\) yields
\(\pi_1p_1^+=\pi_2p_2^-\).
Repeating the same argument recursively gives the local balance relations
\begin{equation}
\label{eq:detailed_balance}
\pi_jp_j^+
=
\pi_{j+1}p_{j+1}^-,
\qquad
j=0,1,\ldots.
\end{equation}
Equation~\eqref{eq:detailed_balance} gives the recursion
\(\pi_{j+1}=\pi_j p_j^+/p_{j+1}^-\), from which the stationary probabilities have the product form
\begin{equation}
\label{eq:stationary_product}
\pi_j
=
\pi_0
\prod_{k=0}^{j-1}
\frac{p_k^+}{p_{k+1}^-},
\qquad
j > 0.
\end{equation}
\subsection{Existence and Uniqueness of a Stationary Distribution}
\label{subsec:stationary_existence}
\begin{restatable}[Existence and uniqueness of a stationary distribution]{mytheorem}
{stationaryexistence}
\label{thm:stationary_existence}
For any fixed \(\varepsilon>0\), both birth--death Markov chains induced by the transition probabilities in \eqref{eq:transition_original} and \eqref{eq:transition_modified} admit a unique stationary distribution on the geometric grid \(j=0,1,\ldots\).
\end{restatable}

The proof is given in Appendix~\ref{app:proofs}.
Theorem~\ref{thm:stationary_existence} verifies the standard product-form normalization condition for birth--death chains.
In the notation of \eqref{eq:stationary_product}, this condition is the convergence of the series
\begin{equation*}
\sum_{j=1}^{\infty}
\prod_{k=0}^{j-1}
\frac{p_k^+}{p_{k+1}^-}
<\infty .
\end{equation*}
This is the discrete birth--death analogue of the classical Karlin--McGregor normalization condition \citep[Ch.~7, Sec.~5]{Karlin1968FirstCourse}.
The proof also uses Proposition~\ref{prop:signed_gap_variance_asymptotics} to show that the noise scales \(s_L(T_j)\) and \(s_R(T_j)\) vanish as \(T_j\to\infty\), and Proposition~\ref{prop:plateau_pass_probability_asymptotics} to convert this variance decay into \(\alpha_{L,j}\to1\) and \(\alpha_{R,j}\to1\).
Consequently, at large ensemble sizes the chain shifts left with probability tending to one and shifts right with probability tending to zero.

Under the folded-normal transition model, the neighboring transition probabilities are strictly positive, so the birth--death chain is irreducible on the geometric grid.
Irreducibility implies that the stationary distribution, once it exists, is unique.
Moreover, for an irreducible countable-state Markov chain, the existence of a stationary distribution is equivalent to positive recurrence \citep[see, e.g., Sec.~1.7]{Norris1998Markov}.
In practical terms, positive recurrence means that the chain returns to its recurrent states in finite expected time; hence fluctuations away from the typical stationary region are not transient excursions to infinity.
Thus, the stationary regime is not only formally normalizable, but also dynamically stable.
\subsection{Equilibrium Level}
\label{subsec:equilibrium_level}
\begin{figure}[t]
\centering
\includegraphics[width=0.95\linewidth]{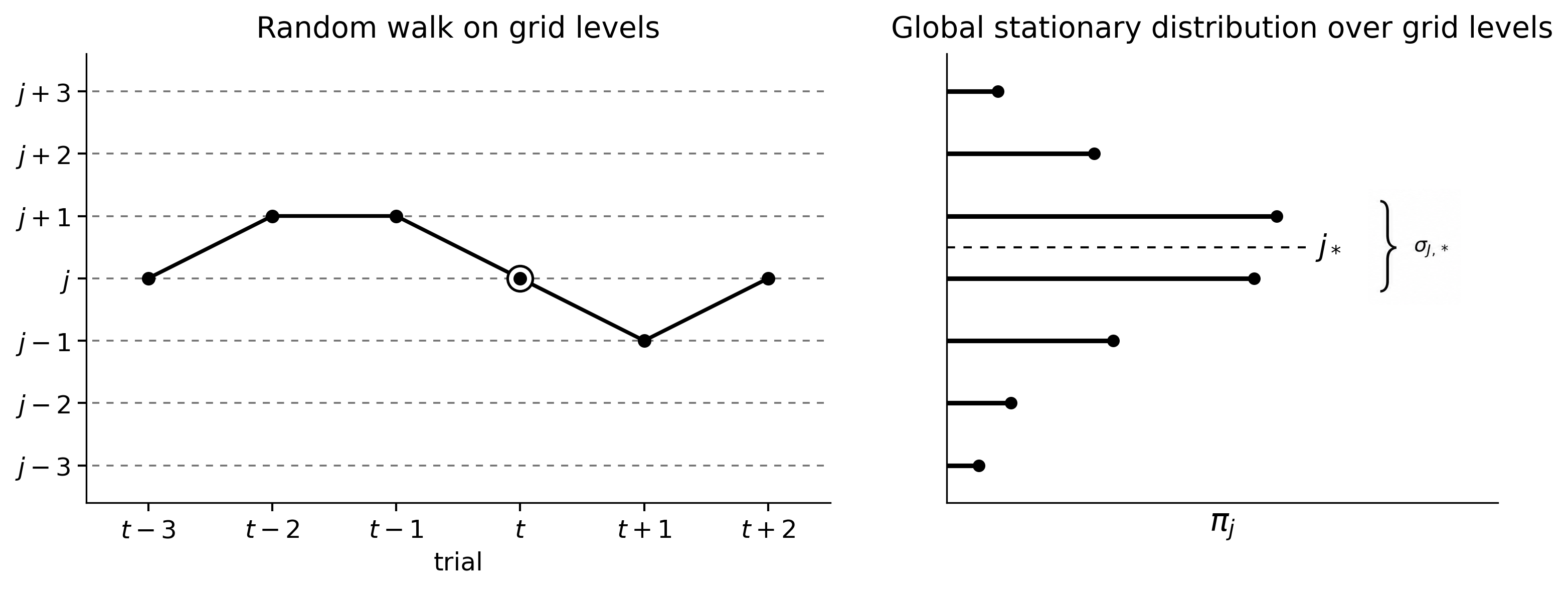}
\caption{Grid-level random walk and schematic stationary masses \(\pi_j\) centered near \(j_*\).}
\label{fig:pi_distribution}
\end{figure}

The stationary distribution is expected to concentrate near the grid levels where the local drift changes sign.
Figure~\ref{fig:pi_distribution} illustrates the transition from a realized trajectory of the grid-level process to its stationary distribution.
The equilibrium center \(j_*\) need not coincide with an integer grid level; it is interpreted as the center of the local Gaussian approximation to the stationary masses \(\pi_j\).
In the continuous approximation, the equilibrium level \(j_*\) is determined by equality of the right- and left-shift probabilities,
\begin{equation}
\label{eq:equilibrium_general}
p_{j_*}^+
\approx
p_{j_*}^- .
\end{equation}
Equation~\eqref{eq:equilibrium_general} is the zero-drift condition that identifies the center of the stationary regime in the continuous approximation.

For compactness, write \(F_0(y)=2\Phi(y)-1\) and \(y_j=\varepsilon/s_R(T_j)\).
It follows from \eqref{eq:left_gap_variance}--\eqref{eq:right_gap_variance} that
\(s_L(T_j)\sim\sqrt{\mathrm{sf}}\,s_R(T_j)\).
Therefore, using the leading centered folded-normal approximation \eqref{eq:alpha_leading},
\[
\alpha_{L,j}
\approx
F_0\left(\frac{y_j}{\sqrt{\mathrm{sf}}}\right),
\qquad
\alpha_{R,j}
\approx
F_0(y_j).
\]

For the original update rule, substituting \eqref{eq:transition_original} into
\eqref{eq:equilibrium_general} gives
\(1-\alpha_{R,j_*}=\alpha_{L,j_*}\alpha_{R,j_*}\).
Hence \(y_*^{\mathrm{orig}}=y_{j_*}\) satisfies
\begin{equation}
\label{eq:balance_original}
F_0(y_*^{\mathrm{orig}})
\left[
1+
F_0\left(\frac{y_*^{\mathrm{orig}}}{\sqrt{\mathrm{sf}}}\right)
\right]
=
1 .
\end{equation}
For the modified update rule, substituting \eqref{eq:transition_modified} into
\eqref{eq:equilibrium_general} gives
\[
\left[
1-F_0\left(\frac{y_*^{\mathrm{mod}}}{\sqrt{\mathrm{sf}}}\right)
\right]
\left[
1-F_0(y_*^{\mathrm{mod}})
\right]
=
F_0\left(\frac{y_*^{\mathrm{mod}}}{\sqrt{\mathrm{sf}}}\right)
F_0(y_*^{\mathrm{mod}}).
\]
Equivalently, this reduces to the symmetric balance equation
\begin{equation}
\label{eq:balance_modified_symmetric}
F_0\left(\frac{y_*^{\mathrm{mod}}}{\sqrt{\mathrm{sf}}}\right)
+
F_0(y_*^{\mathrm{mod}})
=
1 .
\end{equation}
This simplification is a direct consequence of assigning the mixed case, in which one plateau test passes and the other fails, to the stay decision.
Since \(F_0\) is strictly increasing from \(0\) to \(1\), both \eqref{eq:balance_original} and \eqref{eq:balance_modified_symmetric} have unique positive solutions for each fixed \(\mathrm{sf}>1\).

The balance equation \eqref{eq:balance_modified_symmetric} has a simple interpretation.
Since mixed outcomes are assigned to the stay decision, only the two joint events ``both tests fail'' and ``both tests pass'' contribute to the drift.
Thus, equilibrium requires these two probabilities to be equal:
\((1-\alpha_L)(1-\alpha_R)=\alpha_L\alpha_R\), which reduces to
\(\alpha_L+\alpha_R=1\).
Because the left gap has the larger variance, typically \(\alpha_L<\alpha_R\).
Together with \(\alpha_L+\alpha_R=1\), this gives \(\alpha_L<1/2<\alpha_R\).

Once \(y_*^{\mathrm{orig}}\) or \(y_*^{\mathrm{mod}}\) is found from
\eqref{eq:balance_original} or \eqref{eq:balance_modified_symmetric}, respectively, the corresponding equilibrium ensemble size follows from \(y_*=\varepsilon/s_R(B_*)\), where \(B_*=T_0\cdot\mathrm{sf}^{j_*}\) denotes the tree count at the equilibrium level.
Using \eqref{eq:right_gap_variance}, this gives
\begin{equation}
\label{eq:B_star_variants}
B_*^{\mathrm{orig}}
\approx
\frac{v(1-\mathrm{sf}^{-1})}{S_\infty^2}
\frac{\left[y_*^{\mathrm{orig}}(\mathrm{sf})\right]^2}{\varepsilon^2},
\qquad
B_*^{\mathrm{mod}}
\approx
\frac{v(1-\mathrm{sf}^{-1})}{S_\infty^2}
\frac{\left[y_*^{\mathrm{mod}}(\mathrm{sf})\right]^2}{\varepsilon^2}.
\end{equation}

Since the \(y_*\)-constants are independent of \(\varepsilon\), the preceding derivation yields the following scaling result.
\begin{restatable}[Equilibrium scaling with the tolerance]{myproposition}{equilibriumscaling}
\label{prop:equilibrium_scaling}
Under the leading centered folded-normal approximation \eqref{eq:alpha_leading}, for any fixed scale factor \(\mathrm{sf}>1\), the equilibrium ensemble sizes satisfy
\(B_*^{\mathrm{orig}}=O(\varepsilon^{-2})\) and
\(B_*^{\mathrm{mod}}=O(\varepsilon^{-2})\) as \(\varepsilon\downarrow0\), with the leading constants given in \eqref{eq:B_star_variants}.
\end{restatable}
Proposition~\ref{prop:equilibrium_scaling} has a direct practical interpretation.
The tolerance \(\varepsilon\) controls the equilibrium ensemble size quadratically: decreasing \(\varepsilon\) by a factor of two increases the equilibrium number of trees by approximately a factor of four, all else being equal.

Moreover, since the constants \(y_*^{\mathrm{orig}}(\mathrm{sf})\) and \(y_*^{\mathrm{mod}}(\mathrm{sf})\) depend only on the scale factor and the update rule, they can be precomputed once for commonly used values such as \(\mathrm{sf}=1.5\) or \(\mathrm{sf}=2.0\).
The formulas \eqref{eq:B_star_variants} still contain the unknown problem-dependent factor \(v/S_\infty^2\).
This does not prevent their use: the result identifies the functional dependence on \(\varepsilon\) and \(\mathrm{sf}\) up to a single scale coefficient.
This coefficient may later be treated as a nuisance parameter in trajectory-based calibration, but such estimation is outside the scope of the present paper.
\subsection{Original Versus Modified Equilibrium}
\label{subsec:original_vs_modified_equilibrium}
The two update rules differ only in the mixed case \(d_{L,t}\le\varepsilon\), \(d_{R,t}>\varepsilon\).
The original rule assigns this case to a right shift, whereas the modified rule assigns it to staying at the current level.
Therefore, relative to the modified rule, the original rule has a stronger preference for right shifts and should be expected to place the stationary regime at a larger ensemble size.

This effect can be quantified directly from \eqref{eq:B_star_variants}.
The unknown factor \(v(1-\mathrm{sf}^{-1})/(S_\infty^2\varepsilon^2)\) cancels in the ratio, giving
\begin{equation}
\label{eq:orig_mod_ratio}
\frac{B_*^{\mathrm{orig}}}{B_*^{\mathrm{mod}}}
\approx
\left(
\frac{y_*^{\mathrm{orig}}(\mathrm{sf})}
     {y_*^{\mathrm{mod}}(\mathrm{sf})}
\right)^2 .
\end{equation}
Thus, the relative inflation of the equilibrium ensemble size caused by the original mixed-case assignment depends only on the scale factor \(\mathrm{sf}\), and not on \(\varepsilon\) or on the problem-dependent factor \(v/S_\infty^2\).

The limiting case \(\mathrm{sf}\to1^+\) is available analytically.
Since \(F_0(y/\sqrt{\mathrm{sf}})\to F_0(y)\), the modified balance equation \eqref{eq:balance_modified_symmetric} gives \(2F_0(y)=1\), hence
\(y_*^{\mathrm{mod}}(1^+)=\Phi^{-1}(3/4)\).
For the original rule, \eqref{eq:balance_original} gives
\(F_0(y)(1+F_0(y))=1\), and therefore
\(F_0(y)=(\sqrt{5}-1)/2\).
Consequently,
\[
y_*^{\mathrm{orig}}(1^+)
=
\Phi^{-1}\left(\frac{1+\sqrt{5}}{4}\right),
\qquad
y_*^{\mathrm{mod}}(1^+)
=
\Phi^{-1}\left(\frac{3}{4}\right),
\]
and taking the ratio gives
\[
R_0=\lim_{\mathrm{sf}\to1^+}
\frac{B_*^{\mathrm{orig}}}{B_*^{\mathrm{mod}}}
=
\left[
\frac{
\Phi^{-1}\left((1+\sqrt{5})/4\right)
}{
\Phi^{-1}(3/4)
}
\right]^2
\approx
1.68 .
\]
Moreover, expanding the implicit equations around \(\mathrm{sf}=1\) gives
\[
\frac{B_*^{\mathrm{orig}}}{B_*^{\mathrm{mod}}}
=
R_0
\left[
1-\frac{\mathrm{sf}-1}{2\sqrt{5}}
+
O\left((\mathrm{sf}-1)^2\right)
\right].
\]

\begin{figure}[t]
\centering
\includegraphics[width=.85\linewidth]{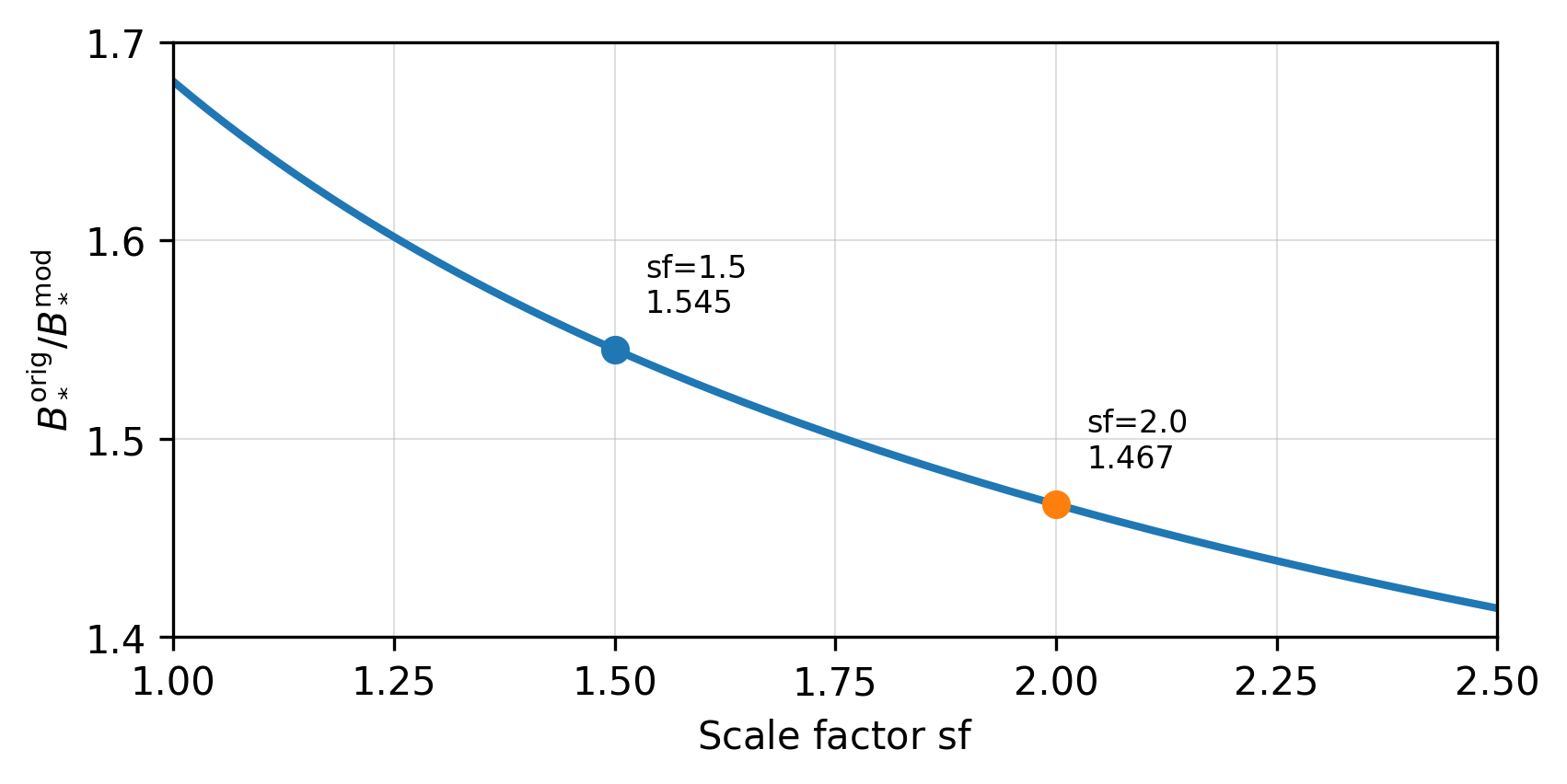}
\caption{Relative inflation of the equilibrium ensemble size induced by the original mixed-case assignment, measured by \(B_*^{\mathrm{orig}}/B_*^{\mathrm{mod}}\), as a function of the scale factor \(\mathrm{sf}\). 
The constants \(y_*^{\mathrm{orig}}(\mathrm{sf})\) and \(y_*^{\mathrm{mod}}(\mathrm{sf})\) are obtained by solving \eqref{eq:balance_original} and \eqref{eq:balance_modified_symmetric}, respectively. 
The points \(\mathrm{sf}=1.5\) and \(\mathrm{sf}=2.0\) are highlighted.}
\label{fig:orig_mod_ratio_sf}
\end{figure}

Thus, near the fine-grid limit \(\mathrm{sf}=1\), the original rule inflates the equilibrium ensemble size by about \(68\%\) relative to the symmetric modified rule, and this inflation decreases as \(\mathrm{sf}\) moves away from one.
Figure~\ref{fig:orig_mod_ratio_sf} shows the numerical solution of \eqref{eq:balance_original} and \eqref{eq:balance_modified_symmetric} over \(\mathrm{sf}\in(1,2.5]\), with the commonly used values \(\mathrm{sf}=1.5\) and \(\mathrm{sf}=2.0\) highlighted.
For \(\mathrm{sf}=1.5\), the ratio in \eqref{eq:orig_mod_ratio} is approximately \(1.545\), meaning that the original update rule places the equilibrium ensemble size about \(54.5\%\) higher than the modified rule.
For \(\mathrm{sf}=2.0\), the corresponding ratio is approximately \(1.467\), or about \(46.7\%\) higher.

From the perspective of HPO efficiency, this creates a trade-off.
The previous experiments in \citet{Porvatov2026How} showed that using \(\mathrm{sf}=1.5\) can be statistically significantly faster than the classical doubling strategy of \citet{Oshiro2012How}, \(\mathrm{sf}=2.0\), because the procedure spends less time training forests corresponding to the expensive right endpoints \(R_t=B_t\cdot\mathrm{sf}\).
At the same time, the relative inflation induced by the conservative original rule is larger for \(\mathrm{sf}=1.5\) than for \(\mathrm{sf}=2.0\).
Thus, smaller scale factors lower the cost of individual right probes, but the conservative mixed-case assignment raises the equilibrium level relative to the symmetric modified benchmark.

This comparison also clarifies the role of the triplet-based plateau mechanism relative to conventional early-stopping heuristics.
Early-stopping methods typically scan ensemble sizes from left to right and stop once a local improvement criterion falls below a tolerance, for example when the left-gap condition \(d_{L,t}\le\varepsilon\) is first satisfied.
Such one-sided stopping can lead to systematic underestimation of the sufficient number of trees.
In contrast, the plateau algorithm performs stochastic updates around an equilibrium region: erroneous left and right corrections can compensate over trials.
The original rule was designed as a more conservative variant by assigning the mixed case \(d_{L,t}\le\varepsilon<d_{R,t}\) to a right shift.
The analysis above shows that this conservatism has a measurable cost: relative to the symmetric modified rule, it shifts the stationary ensemble size upward by a factor depending only on \(\mathrm{sf}\).
\subsection{Local Gaussian Approximation of the Stationary Distribution}
\label{subsec:local_stationary_approximation}
The product-form representation \eqref{eq:stationary_product} gives the stationary distribution exactly, up to normalization.
However, for interpretation and estimation it is useful to approximate this distribution locally near its equilibrium center.
This subsection derives such an approximation directly from the discrete local-balance relation.

Let \(J_t\) denote the grid-level process, so that \(B_t=T_{J_t}=T_0\cdot\mathrm{sf}^{J_t}\).
From \eqref{eq:detailed_balance}, the adjacent stationary probabilities satisfy
\begin{equation}
\label{eq:local_log_ratio}
\log\frac{\pi_{j+1}}{\pi_j}
=
g_j,
\qquad
g_j
=
\log\frac{p_j^+}{p_{j+1}^-}.
\end{equation}
Thus, the local shape of the stationary distribution is controlled by the log-ratio \(g_j\).
If \(g_j>0\), the stationary mass increases from level \(j\) to level \(j+1\); if \(g_j<0\), it decreases.
Therefore, the center of the stationary region is located near the point where this log-ratio changes sign.

For a local continuous approximation, regard \(g_j\) as the value of a smooth function \(g(u)\) at the midpoint \(u=j+1/2\) between the adjacent grid levels \(j\) and \(j+1\).
This midpoint convention is natural because \(\log(\pi_{j+1}/\pi_j)\) compares the masses on the two sides of the midpoint \(j+1/2\).
Let \(j_*\) be the local equilibrium center, defined by \(g(j_*)=0\).
Around this point, \(\pi_j\) is expected to have a local maximum: to the left of \(j_*\) the mass increases with \(j\), while to the right of \(j_*\) it decreases.
Equivalently, \(g(u)\) crosses zero with negative slope.
Thus, near \(j_*\), write \(g(u)\approx g'(j_*)(u-j_*)\), where \(g'(j_*)<0\) under inward drift.

Now compare this with a Gaussian approximation on the grid-level scale, \(\pi_j\propto\exp\{-(j-j_*)^2/(2\sigma_{J,*}^2)\}\).
For such a Gaussian sequence,
\[
\log\frac{\pi_{j+1}}{\pi_j}
=
-\frac{j+1/2-j_*}{\sigma_{J,*}^2}.
\]
Matching this expression with the linear expansion of \(g(u)\) at \(u=j+1/2\) yields
\begin{equation}
\label{eq:sigma_J_local_balance}
\sigma_{J,*}^2
\approx
-\frac{1}{g'(j_*)}.
\end{equation}
The sign condition \(g'(j_*)<0\) guarantees that the variance in \eqref{eq:sigma_J_local_balance} is positive.

The approximation in \eqref{eq:sigma_J_local_balance} is discrete in origin.
It comes from the exact local-balance relation \eqref{eq:local_log_ratio}; the only approximation is the local linearization of \(g(u)\), or equivalently the quadratic approximation of \(\log\pi_j\) near its maximum.
In practice, \(g'(j_*)\) can be evaluated from the transition probabilities \eqref{eq:transition_original} or \eqref{eq:transition_modified}, using the plateau-pass probabilities \(\alpha_{L,j}\) and \(\alpha_{R,j}\).
Equivalently, it can be estimated numerically from a local fit of \(g_j\) around the equilibrium region.

This also gives a regression interpretation of the Gaussian approximation.
Over a window of grid levels near the stationary center, one may fit \(g_j\approx\beta_0+\beta_1(j+1/2)\).
Comparing this with \(g_j\approx-(j+1/2-j_*)/\sigma_{J,*}^2\) gives
\begin{equation*}
\sigma_{J,*}^2\approx-\frac{1}{\beta_1},
\qquad
j_*\approx-\frac{\beta_0}{\beta_1}.
\end{equation*}
This is a least-squares approximation to the local-balance equations, rather than a likelihood model for the observed HPO trajectory.

Finally, the grid-level approximation can be translated to the original ensemble-size scale.
Since \(B=T_0\cdot\mathrm{sf}^J\), the derivative with respect to the grid level is \(dB/dJ=B\log(\mathrm{sf})\).
At the equilibrium point \(B_*=T_0\cdot\mathrm{sf}^{j_*}\), the first-order delta-method conversion gives
\begin{equation}
\label{eq:sigma_B_delta}
\sigma_{B,*}
\approx
\left.
\frac{d}{dJ}
\left(T_0\cdot\mathrm{sf}^{J}\right)
\right|_{J=j_*}
\sigma_{J,*}
\approx
B_*\log(\mathrm{sf})\,\sigma_{J,*}.
\end{equation}
Thus, even a moderate spread on the grid-level scale can correspond to a large absolute spread in the number of trees when \(B_*\) is large.
This \(B\)-scale fluctuation should not be confused with the relative OOB score-gap fluctuations used in the plateau tests; the latter enter the pass probabilities through the variance asymptotics of Proposition~\ref{prop:signed_gap_variance_asymptotics}.
\subsection{Fokker--Planck Interpretation}
The local variance formula \eqref{eq:sigma_J_local_balance} also has a diffusion interpretation.
For one step of the grid-level chain, let \(\Delta J_t=J_{t+1}-J_t\).
Then
\begin{equation*}
\mathbb{E}[\Delta J_t\mid J_t=j]=p_j^+-p_j^-,
\qquad
\mathbb{E}[(\Delta J_t)^2\mid J_t=j]=p_j^++p_j^- .
\end{equation*}
Thus, in a continuous approximation,
\(a(j)=p_j^+-p_j^-\) plays the role of the drift coefficient, while
\(b(j)=p_j^++p_j^-\) plays the role of the one-step second-moment coefficient.
The condition \(p_{j_*}^+=p_{j_*}^-\) is the zero-drift condition
\(\mathbb{E}[\Delta J_t\mid J_t=j_*]=a(j_*)=0\), which defines the equilibrium point \(j_*\).
The conditional variance of one step is
\(\operatorname{Var}(\Delta J_t\mid J_t=j)=b(j)-a^2(j)\).
At \(j_*\), however, \(a(j_*)=0\), so the second moment \(b(j_*)\) and the conditional variance coincide.

The corresponding stationary Fokker--Planck equation has the formal zero-flux form
\begin{equation*}
a(j)\pi(j)
-
\frac{1}{2}\frac{d}{dj}\{b(j)\pi(j)\}
=
0.
\end{equation*}
Linearizing near the equilibrium point, where \(a(j_*)=0\), gives
\(a(j)\approx a'(j_*)(j-j_*)\) and \(b(j)\approx b(j_*)\).
This is a local Ornstein--Uhlenbeck-type approximation: the restoring drift is linearized, while the local step activity is treated as constant over the neighborhood in which the stationary mass is concentrated.
Substituting these approximations into the zero-flux equation yields
\begin{equation*}
a'(j_*)(j-j_*)\pi(j)
-
\frac{b(j_*)}{2}\pi'(j)
=
0.
\end{equation*}
Hence \(\pi'(j)/\pi(j)=2a'(j_*)(j-j_*)/b(j_*)\), and integration gives
\begin{equation*}
\pi(j)
\propto
\exp\left\{
\frac{a'(j_*)}{b(j_*)}(j-j_*)^2
\right\}.
\end{equation*}
Since inward drift implies \(a'(j_*)<0\), this is a Gaussian density with local variance
\begin{equation}
\label{eq:fokker_planck_variance}
\sigma_{J,*}^2
\approx
\frac{b(j_*)}{-2a'(j_*)}.
\end{equation}

To relate this expression to \eqref{eq:sigma_J_local_balance}, consider a local continuous analogue of \eqref{eq:local_log_ratio}.
Instead of the discrete neighboring ratio \(p_j^+/p_{j+1}^-\), write
\(h(j)=\log[p^+(j)/p^-(j)]\), where \(p^+(j)\) and \(p^-(j)\) are smooth local versions of the right- and left-shift probabilities.
At the equilibrium point \(j_*\), the zero-drift condition gives
\(p^+(j_*)=p^-(j_*)=:p_*\).
Therefore,
\begin{equation*}
h'(j_*)
=
\frac{(p^+)'(j_*)}{p^+(j_*)}
-
\frac{(p^-)'(j_*)}{p^-(j_*)}
=
\frac{(p^+)'(j_*)-(p^-)'(j_*)}{p_*}
=
\frac{a'(j_*)}{p_*}.
\end{equation*}
Since \(b(j_*)=p^+(j_*)+p^-(j_*)=2p_*\), it follows that
\(h'(j_*)=2a'(j_*)/b(j_*)\).
Thus the local-balance variance \(\sigma_{J,*}^2\approx-1/h'(j_*)\) reduces to the Fokker--Planck expression \eqref{eq:fokker_planck_variance}.
The only difference is that the exact discrete local-balance relation in \eqref{eq:local_log_ratio} uses the neighboring denominator \(p_{j+1}^-\), whereas the Fokker--Planck argument replaces it by the local smooth value \(p^-(j)\).

The Fokker--Planck argument is therefore best viewed as an interpretation of the discrete result rather than as its replacement.
It introduces an additional continuum approximation, whereas \eqref{eq:sigma_J_local_balance} is obtained directly from the product-form local balance of the birth--death chain.
In the next subsection, \eqref{eq:fokker_planck_variance} is used rather than \eqref{eq:sigma_J_local_balance}, since it separates the local drift term \(a(j)\) from the one-step fluctuation term \(b(j)\).
\subsection{Evaluating the Stationary Variance}
\label{subsec:evaluating_stationary_variance}
The local Gaussian approximation in \eqref{eq:sigma_J_local_balance} reduces the problem of estimating the stationary spread to the evaluation of the local slope \(g'(j_*)\).
This slope can be evaluated analytically under the leading folded-normal transition model, or estimated numerically from the local behavior of the log-ratio \(g_j\) around the stationary region.

The exact discrete expression in \eqref{eq:local_log_ratio} uses the neighboring denominator \(p_{j+1}^-\).
For the local calculation below, this neighboring term is replaced by its local smooth value, and
\(h(j)=\log[p^+(j)/p^-(j)]\)
is used.
This replacement is consistent with the zero-drift approximation \(p^+(j_*)\approx p^-(j_*)\).
Evaluating \(h'(j)\) requires differentiating the pass probabilities with respect to the grid level.
To make this operation well defined, the auxiliary birth--death chain is considered in which the leading approximation \eqref{eq:alpha_leading} is treated as an equality, rather than as an asymptotic approximation.
For this auxiliary chain, with \(B_j=T_0\cdot\mathrm{sf}^j\), the transition probabilities are smooth functions of the continuous variable \(j\).

Under this leading model, \(y_j=\varepsilon/s_R(B_j)\) has the form
\begin{equation*}
y_j
=
C\cdot\mathrm{sf}^{j/2},
\qquad
C
=
\varepsilon \sqrt{\frac{ S_\infty^2 T_0}{v(1-\mathrm{sf}^{-1})}}.
\end{equation*}
Hence \(dy_j/dj=\log(\mathrm{sf})\,y_j/2\), so the constant \(C\), and therefore the factors \(\varepsilon\), \(S_\infty\), \(v\), and \(T_0\), drop out of the local slope calculation.
Consequently, the stationary variance constants below depend only on \(\mathrm{sf}\) and on the update rule.
\begin{restatable}[Stationary variance constants]{myproposition}{stationaryvarianceconstants}
\label{prop:stationary_variance_constants}
Consider the auxiliary birth--death chain on the grid \(B_j=T_0\cdot\mathrm{sf}^j\) in which the leading centered folded-normal approximation \eqref{eq:alpha_leading} is used as the exact transition model.
Then, for any fixed scale factor \(\mathrm{sf}>1\), the zero-drift condition
\eqref{eq:equilibrium_general}, together with the local variance formula
\eqref{eq:fokker_planck_variance}, gives the following approximations for the
grid-level stationary variance constants
\begin{align}
\label{eq:sigma_J_orig}
\sigma_{J,*}^{2,\mathrm{orig}}
&\approx
\frac{
F_0\left(y_*^{\mathrm{orig}}\right)
\left[1-F_0\left(y_*^{\mathrm{orig}}\right)\right]
}
{
\log(\mathrm{sf})\,y_*^{\mathrm{orig}}
\left[
\phi\left(y_*^{\mathrm{orig}}\right)
+
\dfrac{
\phi\left(y_*^{\mathrm{orig}}/\sqrt{\mathrm{sf}}\right)
}
{\sqrt{\mathrm{sf}}}
F_0^2\left(y_*^{\mathrm{orig}}\right)
\right]
},
\\[1mm]
\label{eq:sigma_J_mod}
\sigma_{J,*}^{2,\mathrm{mod}}
&\approx
\frac{
F_0(y_*^{\mathrm{mod}})
F_0(y_*^{\mathrm{mod}}/\sqrt{\mathrm{sf}})
}
{
\log(\mathrm{sf})\,y_*^{\mathrm{mod}}
\left[
\phi(y_*^{\mathrm{mod}})
+
\dfrac{\phi(y_*^{\mathrm{mod}}/\sqrt{\mathrm{sf}})}{\sqrt{\mathrm{sf}}}
\right]
}.
\end{align}
Here \(y_*^{\mathrm{orig}}\) and \(y_*^{\mathrm{mod}}\) are defined by
\eqref{eq:balance_original} and \eqref{eq:balance_modified_symmetric}, respectively.
\end{restatable}

The proof is given in Appendix~\ref{app:proofs}.
The result uses the transition probabilities \eqref{eq:transition_original} and \eqref{eq:transition_modified}, together with the leading plateau-pass approximation \eqref{eq:alpha_leading}.
The proof differentiates the drift \(a(j)=p^+(j)-p^-(j)\), uses
\(dy_j/dj=\log(\mathrm{sf})\,y_j/2\), and then substitutes the balance equations
\eqref{eq:balance_original} and \eqref{eq:balance_modified_symmetric} into the Fokker--Planck variance formula \eqref{eq:fokker_planck_variance}.
Proposition~\ref{prop:stationary_variance_constants} shows that, in the leading model, the grid-level stationary variance constants in \eqref{eq:sigma_J_orig} and \eqref{eq:sigma_J_mod} are independent of the tolerance \(\varepsilon\) and depend only on \(\mathrm{sf}\) and on the update rule.
Therefore, similarly to the equilibrium constants in \eqref{eq:B_star_variants}, they can be precomputed for commonly used scale factors.
Changing \(\varepsilon\) shifts the equilibrium location but does not change the local variance constant on the \(J\)-scale.

For the modified rule, \eqref{eq:sigma_J_mod} has a useful physical interpretation.
It is derived from the Fokker--Planck variance formula
\eqref{eq:fokker_planck_variance}.
At the modified equilibrium, \eqref{eq:balance_modified_symmetric} gives
\(\alpha_L(j_*)+\alpha_R(j_*)=1\).
Together with the modified transition probabilities in \eqref{eq:transition_modified}, this implies
\[
p_*
=
p_{j_*}^+
=
p_{j_*}^-
=
\alpha_L(j_*)\alpha_R(j_*)
=
[1-\alpha_L(j_*)][1-\alpha_R(j_*)].
\]
Thus \(p_*\) is the common probability of a non-stay move to the right or to the left at equilibrium.
Moreover, \eqref{eq:transition_modified} gives
\(a(j)=p^+(j)-p^-(j)=1-\alpha_L(j)-\alpha_R(j)\).
Therefore,
\(-a'(j_*)=\left.d(\alpha_L+\alpha_R)/dj\right|_{j=j_*}\).
Since the one-step second-moment coefficient at equilibrium is
\(b(j_*)=p_{j_*}^+ + p_{j_*}^- = 2p_*\), the Fokker--Planck expression
\eqref{eq:fokker_planck_variance} becomes
\[
\sigma_{J,*}^{2,\mathrm{mod}}
\approx
\frac{p_*}
{
\left.
\dfrac{d}{dj}\bigl(\alpha_L(j)+\alpha_R(j)\bigr)
\right|_{j=j_*}
}.
\]
Hence the stationary variance is proportional to the local probability of actual shifts, rather than stay decisions, and inversely proportional to the combined sensitivity of the left and right plateau-pass probabilities to perturbations of the equilibrium level.
In the local-balance view, this is the same mechanism as
\(\sigma_{J,*}^2\approx-1/g'(j_*)\): a sharper peak of the stationary distribution corresponds to a smaller variance.

The grid-level variance can be converted to the ensemble-size scale by \eqref{eq:sigma_B_delta}.
For the original and modified update rules,
\begin{equation}
\label{eq:sigma_B_variants}
\sigma_{B,*}^{\mathrm{orig}}
\approx
B_*^{\mathrm{orig}}\log(\mathrm{sf})\,\sigma_{J,*}^{\mathrm{orig}},
\qquad
\sigma_{B,*}^{\mathrm{mod}}
\approx
B_*^{\mathrm{mod}}\log(\mathrm{sf})\,\sigma_{J,*}^{\mathrm{mod}}.
\end{equation}
Combining the delta-method conversion \eqref{eq:sigma_B_variants} with the equilibrium-size formulas \eqref{eq:B_star_variants} gives the following scaling result.
\begin{restatable}[Stationary-spread scaling with the tolerance]{myproposition}{stationaryspreadbscale}
\label{prop:stationary_spread_B_scale}
Under the leading centered folded-normal approximation \eqref{eq:alpha_leading} and the local variance formula \eqref{eq:fokker_planck_variance}, 
for any fixed scale factor \(\mathrm{sf}>1\), 
the stationary standard deviations on the 
ensemble-size scale satisfy, as \(\varepsilon\downarrow0\),
\begin{align}
\label{eq:sigma_B_orig_scaling}
\sigma_{B,*}^{\mathrm{orig}}
&\approx
\frac{v(1-\mathrm{sf}^{-1})}{S_\infty^2}
\frac{
\left[y_*^{\mathrm{orig}}(\mathrm{sf})\right]^2
\log(\mathrm{sf})\,\sigma_{J,*}^{\mathrm{orig}}(\mathrm{sf})
}
{\varepsilon^2}
=
O(\varepsilon^{-2}),
\\
\label{eq:sigma_B_mod_scaling}
\sigma_{B,*}^{\mathrm{mod}}
&\approx
\frac{v(1-\mathrm{sf}^{-1})}{S_\infty^2}
\frac{
\left[y_*^{\mathrm{mod}}(\mathrm{sf})\right]^2
\log(\mathrm{sf})\,\sigma_{J,*}^{\mathrm{mod}}(\mathrm{sf})
}
{\varepsilon^2}
=
O(\varepsilon^{-2}).
\end{align}
Consequently, the variances on the ensemble-size scale satisfy
\((\sigma_{B,*}^{\mathrm{orig}})^2=O(\varepsilon^{-4})\) and
\((\sigma_{B,*}^{\mathrm{mod}})^2=O(\varepsilon^{-4})\).
\end{restatable}

The proof is given in Appendix~\ref{app:proofs}.
Although the explicit forms of the grid-level variance constants
\eqref{eq:sigma_J_orig}--\eqref{eq:sigma_J_mod} are not substituted into
\eqref{eq:sigma_B_orig_scaling}--\eqref{eq:sigma_B_mod_scaling}, Proposition~\ref{prop:stationary_variance_constants} is used through the fact that
\(\sigma_{J,*}^{\mathrm{orig}}(\mathrm{sf})\) and
\(\sigma_{J,*}^{\mathrm{mod}}(\mathrm{sf})\) are independent of \(\varepsilon\).

Thus, \(B_*\) and \(\sigma_{B,*}\) have the same order in \(\varepsilon\), whereas the variance on the \(B\)-scale has the squared order.
This is different from a Poisson-type scaling, where the variance is of the same order as the mean.
The difference is caused by the geometric grid: a one-level fluctuation around \(B_*\) corresponds to an absolute change of order \(B_*\) in the number of trees.

The relative leading spread, i.e. the coefficient-of-variation-like quantity on the ensemble-size scale, is
\[
\frac{\sigma_{B,*}}{B_*}
\approx
\log(\mathrm{sf})\,\sigma_{J,*},
\]
which is independent of \(\varepsilon\) in the leading model.
Therefore, decreasing \(\varepsilon\) increases both the equilibrium ensemble size and the absolute spread around it, while preserving the leading relative spread.

Table~\ref{tab:stationary_constants_sf} provides a numerical evaluation of the constants that enter the stationary-center and stationary-spread formulas.
For each fixed value of \(\mathrm{sf}\), the one-dimensional nonlinear balance equations defining \(y_*^{\mathrm{orig}}\) and \(y_*^{\mathrm{mod}}\) are first solved, and the resulting roots are then substituted into \eqref{eq:sigma_J_orig}--\eqref{eq:sigma_J_mod}.
Thus, the table is not an empirical validation on real data sets, but a numerical consequence of the analytical stationary model.
It quantifies the magnitude of the constants that determine the relative spread on the ensemble-size scale.

The value \(\mathrm{sf}=\sqrt{2}\) is included because it corresponds to splitting the classical doubling step \(\mathrm{sf}=2\) into two equal multiplicative substeps.
Although \(\sigma_{J,*}\) decreases as \(\mathrm{sf}\) increases, the relative spread coefficient \(\log(\mathrm{sf})\sigma_{J,*}\) increases.
This is consistent with the geometric parameterization \(B_j=T_0\cdot\mathrm{sf}^j\): larger scale factors make the stationary distribution narrower in grid levels, but each one-level fluctuation corresponds to a larger multiplicative change in the ensemble size.
\begin{table}[t]
\centering
\begin{tabular}{c|cc|cc|cc|cc}
\hline
\(\mathrm{sf}\)
&
\(y_*^{\mathrm{orig}}\)
&
\(y_*^{\mathrm{mod}}\)
&
\(\sigma_{J,*}^{\mathrm{orig}}\)
&
\(\sigma_{J,*}^{\mathrm{mod}}\)
&
\(\log(\mathrm{sf})\,\sigma_{J,*}^{\mathrm{orig}}\)
&
\(\log(\mathrm{sf})\,\sigma_{J,*}^{\mathrm{mod}}\)
&
\(p_*^{\mathrm{orig}}\)
&
\(p_*^{\mathrm{mod}}\)
\\
\hline
\(1.10\) & \(0.886\) & \(0.691\) & \(2.728\) & \(2.473\) & \(0.260\) & \(0.236\) & \(0.376\) & \(0.250\) \\
\(1.25\) & \(0.902\) & \(0.713\) & \(1.769\) & \(1.616\) & \(0.395\) & \(0.361\) & \(0.367\) & \(0.249\) \\
\(\sqrt{2}\) & \(0.918\) & \(0.734\) & \(1.409\) & \(1.296\) & \(0.488\) & \(0.449\) & \(0.359\) & \(0.249\) \\
\(1.50\) & \(0.925\) & \(0.744\) & \(1.299\) & \(1.198\) & \(0.526\) & \(0.486\) & \(0.355\) & \(0.248\) \\
\(2.00\) & \(0.964\) & \(0.796\) & \(0.977\) & \(0.913\) & \(0.677\) & \(0.633\) & \(0.335\) & \(0.245\) \\
\(2.50\) & \(0.994\) & \(0.836\) & \(0.839\) & \(0.792\) & \(0.769\) & \(0.726\) & \(0.320\) & \(0.241\) \\
\hline
\end{tabular}
\caption{Equilibrium constants, grid-level standard deviations, relative spread coefficients, and equilibrium shift probabilities for selected scale factors.}
\label{tab:stationary_constants_sf}
\end{table}

The last two columns also clarify the equilibrium activity of the two update rules.
For the modified rule, \(p_*^{\mathrm{mod}}\) remains close to \(1/4\) over the reported range of scale factors.
Since \(p_*^{\mathrm{mod}}\) is the common left- and right-shift probability at equilibrium, the total probability of a non-stay move is \(2p_*^{\mathrm{mod}}\approx 1/2\).
Thus, in the symmetric modified rule, the process spends roughly half of the stationary steps moving to a neighboring grid level and roughly half staying at the current level.

For the original rule, the common shift probability \(p_*^{\mathrm{orig}}\) is slightly above \(1/3\) for the practically relevant scale factors \(\mathrm{sf}=\sqrt{2}\) and \(\mathrm{sf}=1.5\).
Consequently, the left- and right-shift probabilities are each a little larger than one third, while the stay probability is a little smaller than one third.
This is qualitatively consistent with Figure~\ref{fig:trajectories}, where the central triplet point continues to move frequently in the stationary regime rather than remaining fixed for long stretches.

This also reframes classical stopping criteria based on comparing two neighboring ensemble sizes, for example \(T\) and \(2T\).
In the present notation, when the central point is \(B=2T\) and \(\mathrm{sf}=2\), such a rule corresponds to a single left-gap test \(d_L\le\varepsilon\).
Equivalently, it can be viewed as a right-gap test at the previous grid level, but the left-gap formulation is the one naturally associated with a left-to-right early-stopping scan.
However, Table~\ref{tab:stationary_constants_sf} shows that, for \(\mathrm{sf}=2\), the grid-level stationary standard deviations \(\sigma_{J,*}^{\mathrm{orig}}\) and \(\sigma_{J,*}^{\mathrm{mod}}\) are both already close to one grid step.
For \(\mathrm{sf}=\sqrt{2}\) and \(\mathrm{sf}=1.5\), both standard deviations are even larger than one grid step, although still below one and a half steps.
Thus, the two ensemble sizes involved in an adjacent comparison are separated by a distance comparable to the intrinsic stationary spread of the plateau process.
This places the classical, still widely used adjacent-score stopping criterion under a specific criticism: it should be interpreted as a noisy local diagnostic of plateau behavior rather than as a deterministic stopping certificate.
\subsection{Implications for Stationary Estimation}
\label{subsec:implications_stationary_estimation}
The preceding results identify \(B_*\) and \(\sigma_{B,*}\) as population-level stationary quantities of the plateau process.
This suggests that practical estimators should target the stationary regime rather than a first hitting time or a single best trial.
The construction and benchmarking of such estimators are left to separate work.
\section{Discussion}
\label{sec:discussion}
The analysis in this paper shifts the interpretation of triplet-based plateau search from a stopping mechanism to a stationary stochastic process.
After the non-\(T\) hyperparameters have effectively stabilized, the central triplet point \(B_t\) is not expected to converge to a deterministic ensemble size.
Instead, it fluctuates around a stationary region whose center and spread are determined by the tolerance, the scale factor, and the stochastic behavior of the OOB score gaps.

A first consequence is that the tolerance \(\varepsilon\) has a quadratic effect on the stationary center.
Under the leading centered folded-normal model, both the original and modified update rules satisfy \(B_*=O(\varepsilon^{-2})\).
Thus, decreasing \(\varepsilon\) by a factor of two shifts the equilibrium ensemble size by approximately a factor of four.
The same order also appears for the stationary standard deviation on the ensemble-size scale, \(\sigma_{B,*}=O(\varepsilon^{-2})\).
Consequently, decreasing \(\varepsilon\) increases both the typical number of trees and the absolute magnitude of the stationary fluctuations around it.

This behavior should not be interpreted as a weakness of the plateau procedure.
It reflects the fact that the process evolves on a geometric grid.
A one-level fluctuation near \(B_*\) corresponds to a multiplicative change in the ensemble size, and hence to an absolute change of order \(B_*\).
For this reason, the relative leading spread \(\sigma_{B,*}/B_*\) is independent of \(\varepsilon\) to first order and is controlled mainly by \(\mathrm{sf}\) and by the update rule.

The numerical constants in Table~\ref{tab:stationary_constants_sf} make this effect concrete.
For the scale factor \(\mathrm{sf}=1.5\), used in the plateau experiments illustrated in Figure~\ref{fig:trajectories}, the conservative original rule gives \(\log(\mathrm{sf})\sigma_{J,*}\approx0.526\), while the symmetric modified rule gives \(\log(\mathrm{sf})\sigma_{J,*}\approx0.486\).
Thus, the leading standard deviation on the ensemble-size scale is about one half of the stationary mean in both cases.
This order of magnitude is qualitatively consistent with Figure~\ref{fig:trajectories}, where the stationary fluctuations of the central triplet point are visibly substantial rather than negligible.
A similar one-half order is obtained for the half-doubling grid \(\mathrm{sf}=\sqrt{2}\), where the corresponding constants are \(0.488\) and \(0.449\).
This is a substantial spread, and it becomes even larger under the classical doubling grid \(\mathrm{sf}=2\), where the constants increase to \(0.677\) and \(0.633\), respectively.
This behavior is an intrinsic consequence of using stochastic plateau decisions on a geometric grid rather than a deterministic one-shot stopping threshold.

The comparison between the original and modified update rules also clarifies the role of the mixed case \(d_L\le\varepsilon<d_R\).
The original rule assigns this case to a right shift, making it more conservative and moving the stationary center to larger ensemble sizes.
The modified rule assigns the same case to staying, producing a more symmetric transition structure.
The theory shows that this design choice changes both the equilibrium location and the relative spread by constants depending only on \(\mathrm{sf}\).
Therefore, the choice between the two rules is not merely an implementation detail: it controls a tradeoff between conservative overestimation and a more symmetric stationary regime.

The scale factor \(\mathrm{sf}\) plays a second, distinct role.
Smaller scale factors make each right probe less expensive because the right endpoint \(R_t=B_t\cdot\mathrm{sf}\) is closer to the current central point.
They also reduce the relative spread on the ensemble-size scale.
However, smaller scale factors imply a finer grid and can require more update steps to move across a large range of tree counts.
If \(\mathrm{sf}\) is chosen too close to one, the neighboring ensembles become nearly indistinguishable, and the relative score gaps may become negligible compared with the tolerance.
In that case, both plateau tests tend to pass, and the update rule is biased toward left shifts.
Thus, \(\mathrm{sf}\) should be viewed as a resolution parameter of the stationary search process, rather than only as a multiplicative expansion factor.

The geometric grid also has a scaling justification.
For a general grid with local increment \(\Delta(B)\), the relevant separation between neighboring ensembles is the relative increment \(\Delta(B)/B\).
If \(\Delta(B)=o(B)\), then this relative separation vanishes as \(B\to\infty\), and the neighboring forests become asymptotically indistinguishable on the scale of the plateau tests.
For any fixed tolerance, such a regime makes the signed score gaps increasingly likely to fall inside the plateau interval, producing an artificial tendency toward left shifts.
Thus, a non-degenerate plateau comparison requires \(\Delta(B)\) to be of order \(B\), at least asymptotically.
The geometric grid \(B_j=T_0\cdot\mathrm{sf}^j\) is the simplest construction with this property, since \(B_{j+1}-B_j=(\mathrm{sf}-1)B_j\).

The theory is derived under several idealizations.
It isolates the ensemble-size dynamics after the remaining hyperparameters have stabilized, approximates the signed OOB score gaps by a centered Gaussian model whose absolute values determine the plateau-pass probabilities, and uses an auxiliary birth--death chain in which the leading folded-normal transition probabilities are treated as exact.
These assumptions are useful because they expose the dominant scaling laws and yield closed-form constants, but they should not be interpreted as claiming that finite HPO trajectories follow the limiting model exactly.

These limitations also indicate natural directions for empirical follow-up.
One can test the normal approximation for the signed relative gaps directly, for example using diagnostic plots or normality tests such as Shapiro--Wilk on suitably standardized local gap samples.
One can also examine whether the empirical relative spread \(\sigma_{B,*}/B_*\) remains approximately stable as \(\varepsilon\) varies, as predicted by the leading theory.
Such validation requires repeated trajectories or sufficiently long stationary segments, and is therefore better suited to a separate empirical study focused on finite-sample estimation.

The present work deliberately stops before constructing a full trajectory-based estimator.
The results identify the stationary center \(B_*\) and the stationary spread \(\sigma_{B,*}\) as population-level quantities of the plateau process.
A practical estimator could be based on empirical transition probabilities, local regression of the log-balance ratio, or moment equations derived from the stationary conditions.
The construction and benchmarking of such estimators are left to separate work, where they can be evaluated against early-stopping heuristics and fixed-grid HPO baselines without overloading the theoretical development presented here.
\acks{The work was supported by the grant for research centers in the field of AI provided by the Ministry of Economic Development of the Russian Federation in accordance with the agreement 000000C313925P4F0002 and the agreement with Skoltech No.~139-10-2025-033.
No competing interests are declared.}
\appendix
\section{Proofs}
\label{app:proofs}
\signedgapvarianceasymptotics*
\begin{proof}
The right-gap variance \eqref{eq:right_gap_variance} was derived in \citet{Porvatov2026How}. 
We give the corresponding derivation for the left gap and then indicate how the same calculation recovers the right-gap expression.

Let \(h_L(x,y)=(x-y)/x\), so that \((S_B-S_L)/S_B=h_L(S_B,S_L)\).
The gradient is
$\nabla h_L(x,y)=(y/x^2, -1/x)^\top.$
Applying the first-order delta method to \((S_B,S_L)\mid D\) gives
\begin{align*}
\operatorname{Var}\left[
\frac{S_B-S_L}{S_B}
\;\middle|\;D
\right]
&\approx
\nabla h_L(\mu_B,\mu_L)^\top
\begin{pmatrix}
\sigma_B^2 & \sigma_{BL}\\
\sigma_{BL} & \sigma_L^2
\end{pmatrix}
\nabla h_L(\mu_B,\mu_L)
\\
&=
\begin{pmatrix}
\dfrac{\mu_L}{\mu_B^2} &
-\dfrac{1}{\mu_B}
\end{pmatrix}
\begin{pmatrix}
\sigma_B^2 & \sigma_{BL}\\
\sigma_{BL} & \sigma_L^2
\end{pmatrix}
\begin{pmatrix}
\mu_L/\mu_B^2\\
-1/\mu_B
\end{pmatrix}
\\
&=
\frac{\mu_L^2}{\mu_B^4}\sigma_B^2
+
\frac{1}{\mu_B^2}\sigma_L^2
-
2\frac{\mu_L}{\mu_B^3}\sigma_{BL}.
\end{align*}
Here \(\sigma_B^2=\operatorname{Var}[S_B\mid D]\), \(\sigma_L^2=\operatorname{Var}[S_L\mid D]\), and \(\sigma_{BL}=\operatorname{Cov}[S_B,S_L\mid D]\).
By \eqref{eq:score_convergence}, \(\mu_B,\mu_L\to S_\infty\). 
Using \(L=B/\mathrm{sf}\) and the covariance scaling in \eqref{eq:finite_ensemble_covariance_scaling}, we obtain
\(\sigma_B^2\sim v/B\), \(\sigma_L^2\sim v/L\), and \(\sigma_{BL}\sim v/B\).
Therefore,
\[
\operatorname{Var}\left[
\frac{S_B-S_L}{S_B}
\;\middle|\;D
\right]
\sim
\frac{1}{S_\infty^2}
\left(
\frac{v}{B}
+
\frac{v}{L}
-
2\frac{v}{B}
\right)
=
\frac{v}{S_\infty^2}\frac{\mathrm{sf}-1}{B},
\]
which proves \eqref{eq:left_gap_variance}.

For completeness, the right gap follows from the same calculation with \(h_R(x,y)=(y-x)/x\), applied to \((S_B,S_R)\).
Since $\nabla h_R(x,y)=(-y/x^2, 1/x)^\top$,
the first-order delta-method approximation gives
\[
\operatorname{Var}\left[
\frac{S_R-S_B}{S_B}
\;\middle|\;D
\right]
\sim
\frac{1}{S_\infty^2}
\left(
\frac{v}{B}
+
\frac{v}{R}
-
2\frac{v}{R}
\right)
=
\frac{v}{S_\infty^2}\frac{1-\mathrm{sf}^{-1}}{B},
\]
because \(R=B\cdot\mathrm{sf}\) and \(\sigma_{BR}\sim v/R\).
This recovers \eqref{eq:right_gap_variance}.
\end{proof}
\plateaupassprobabilityasymptotics*
\begin{proof}
We first derive the order of the conditional means of the signed relative gaps.

For the left signed gap, define
\(h_L(x,y)=(x-y)/x\), so that \(h_L(S_B,S_L)=(S_B-S_L)/S_B\).
A second-order Taylor expansion around \((\mu_B,\mu_L)\) gives
\begin{multline*}
h_L(S_B,S_L)
\approx
h_L(\mu_B,\mu_L)
+
\nabla h_L(\mu_B,\mu_L)^\top
\begin{pmatrix}
S_B-\mu_B\\
S_L-\mu_L
\end{pmatrix}
\\
+
\frac12
\begin{pmatrix}
S_B-\mu_B\\
S_L-\mu_L
\end{pmatrix}^{\!\top}
H_{h_L}(\mu_B,\mu_L)
\begin{pmatrix}
S_B-\mu_B\\
S_L-\mu_L
\end{pmatrix},
\end{multline*}
where
\[
\nabla h_L(x,y)
=
\begin{pmatrix}
y/x^2\\
-1/x
\end{pmatrix},
\qquad
H_{h_L}(x,y)
=
\begin{pmatrix}
-2y/x^3 & 1/x^2\\
1/x^2 & 0
\end{pmatrix}.
\]
The zeroth-order term satisfies, by \eqref{eq:score_convergence},
\(h_L(\mu_B,\mu_L)=(\mu_B-\mu_L)/\mu_B=O(B^{-\gamma})\).
Taking conditional expectation, the linear term vanishes and the second-order contribution is
\[
-\frac{\mu_L}{\mu_B^3}\sigma_B^2
+
\frac{1}{\mu_B^2}\sigma_{BL},
\]
where \(\sigma_B^2=\mathrm{Var}[S_B\mid D]\) and
\(\sigma_{BL}=\mathrm{Cov}[S_B,S_L\mid D]\).
By \eqref{eq:finite_ensemble_covariance_scaling}, this contribution is \(O(B^{-1})\).
Therefore
\[
m_L(B)
=
\mathbb{E}\left[
\frac{S_B-S_L}{S_B}
\;\middle|\;D
\right]
=
O(B^{-\gamma})+O(B^{-1})
=
O(B^{-\beta}),
\qquad
\beta=\min\{\gamma,1\}.
\]

For the right signed gap, the same calculation is applied to
\(h_R(x,y)=(y-x)/x\), with \((x,y)=(S_B,S_R)\).
This is exactly the delta-method expansion used in \citet{Porvatov2026How}. 
The zeroth-order term is \(h_R(\mu_B,\mu_R)=(\mu_R-\mu_B)/\mu_B=O(B^{-\gamma})\), while the second-order contribution is
\[
\frac{\mu_R}{\mu_B^3}\sigma_B^2
-
\frac{1}{\mu_B^2}\sigma_{BR}
=
O(B^{-1}).
\]
Hence
\[
m_R(B)
=
\mathbb{E}\left[
\frac{S_R-S_B}{S_B}
\;\middle|\;D
\right]
=
O(B^{-\beta})
\]
with the same \(\beta=\min\{\gamma,1\}\).
Since \(\gamma>1/2\), we have \(\beta>1/2\).

We now turn to the plateau-pass probabilities.
Consider a generic signed gap \(X_B\) with conditional mean \(m(B)\) and standard deviation \(s(B)\). 
Under the Gaussian approximation \(X_B\mid D\sim\mathcal{N}(m(B),s^2(B))\), the folded-normal probability is
\[
\mathbb{P}\left[
|X_B|\le\varepsilon
\;\middle|\;D
\right]
=
\Phi\left(\frac{\varepsilon-m(B)}{s(B)}\right)
-
\Phi\left(\frac{-\varepsilon-m(B)}{s(B)}\right).
\]
Writing \(y(B)=\varepsilon/s(B)\) and \(\delta(B)=m(B)/s(B)\), and using \(\Phi(-x)=1-\Phi(x)\), this becomes
\[
\Phi\left(y(B)-\delta(B)\right)
+
\Phi\left(y(B)+\delta(B)\right)
-1.
\]
Expanding the two terms in powers of \(\delta(B)\) up to the second order gives cancellation of the first-order terms. 
Using \(\phi'(y)=-y\phi(y)\), where \(\phi\) is the standard normal density, we obtain
\[
\Phi(y-\delta)+\Phi(y+\delta)-1
=
2\Phi(y)-1
-
y\phi(y)\delta^2
+
O(\delta^4).
\]
Since \(y\phi(y)\) is bounded for \(y\ge0\), the leading correction
to the centered folded-normal probability is quadratic in the normalized mean, 
i.e. \(O(\delta^2(B))\).

By Proposition~\ref{prop:signed_gap_variance_asymptotics}, \(s_L(B)\asymp B^{-1/2}\) and \(s_R(B)\asymp B^{-1/2}\). 
Together with the mean bounds derived above, this gives
\[
\frac{m_L(B)}{s_L(B)}=O(B^{1/2-\beta}),
\qquad
\frac{m_R(B)}{s_R(B)}=O(B^{1/2-\beta}),
\]
and hence the quadratic correction is \(O(B^{1-2\beta})\).
Applying the generic expansion to the left and right signed gaps yields
\eqref{eq:alpha_left_asymptotic} and \eqref{eq:alpha_right_asymptotic}.
\end{proof}
\stationaryexistence*
\begin{proof}
By \eqref{eq:stationary_product}, any stationary distribution must be proportional to the product-form weights
\[
a_0=1,
\qquad
a_j=
\prod_{k=0}^{j-1}
\frac{p_k^+}{p_{k+1}^-},
\qquad j\ge1 .
\]
For birth--death chains, this product-form construction is standard, and the weights define a stationary probability distribution precisely when the normalizing series is finite \citep[Ch.~7, Sec.~5]{Karlin1968FirstCourse}.
Thus it remains to verify
\begin{equation}
\label{eq:karlin_mcgregor_condition}
1+
\sum_{j=1}^{\infty}
\prod_{k=0}^{j-1}
\frac{p_k^+}{p_{k+1}^-}
<\infty .
\end{equation}

We verify \eqref{eq:karlin_mcgregor_condition} by the ratio test (d'Alembert's criterion). 
Since
\[
\frac{a_{j+1}}{a_j}
=
\frac{p_j^+}{p_{j+1}^-},
\]
it is enough to study the asymptotic behavior of \(p_j^+\) and \(p_{j+1}^-\).
By Proposition~\ref{prop:signed_gap_variance_asymptotics},
\(s_L(T_j)\asymp T_j^{-1/2}\) and \(s_R(T_j)\asymp T_j^{-1/2}\).
Since \(T_j=T_0\cdot\mathrm{sf}^j\to\infty\), for every fixed \(\varepsilon>0\) we have
\(\varepsilon/s_L(T_j)\to\infty\) and \(\varepsilon/s_R(T_j)\to\infty\).
Proposition~\ref{prop:plateau_pass_probability_asymptotics} then gives
\(\alpha_{L,j}\to1\) and \(\alpha_{R,j}\to1\).

For both update rules, the left-shift probability is
\(p_j^-=\alpha_{L,j}\alpha_{R,j}\), hence \(p_j^-\to1\).
For the original update rule, \(p_j^+=1-\alpha_{R,j}\to0\).
For the modified update rule, \(p_j^+=(1-\alpha_{L,j})(1-\alpha_{R,j})\to0\).
Therefore, in both cases,
\[
\frac{a_{j+1}}{a_j}
=
\frac{p_j^+}{p_{j+1}^-}
\to0<1 .
\]
So, by the ratio test, the series in \eqref{eq:karlin_mcgregor_condition} converges.
Thus the product-form weights are normalizable and define a stationary distribution.

Under the folded-normal transition model, the neighboring transition probabilities are strictly positive on the semi-infinite grid, so the birth--death chain is irreducible.
For an irreducible Markov chain, a stationary distribution is unique.
\end{proof}
\stationaryvarianceconstants*
\begin{proof}
\label{app:stationary_variance_constants}
We work with the auxiliary birth--death chain in which the leading plateau-pass approximation \eqref{eq:alpha_leading} is treated as an exact transition model on the grid
\(B_j=T_0\cdot\mathrm{sf}^j\).
As shown before Proposition~\ref{prop:stationary_variance_constants}, this gives
\(dy_j/dj=\log(\mathrm{sf})\,y_j/2\).
Since \(F_0'(y)=2\phi(y)\), it follows that
\[
\frac{d}{dj}F_0(y_j)
=
\log(\mathrm{sf})\,y_j\,\phi(y_j),
\qquad
\frac{d}{dj}F_0\left(y_j/\sqrt{\mathrm{sf}}\right)
=
\log(\mathrm{sf})\,y_j\,
\frac{\phi(y_j/\sqrt{\mathrm{sf}})}{\sqrt{\mathrm{sf}}}.
\]

For the original update rule, \eqref{eq:transition_original} gives
\(p_j^+=1-F_0(y_j)\) and
\(p_j^-=F_0(y_j/\sqrt{\mathrm{sf}})F_0(y_j)\).
Thus
\[
a(j)=p_j^+-p_j^-
=
1-F_0(y_j)
-
F_0(y_j/\sqrt{\mathrm{sf}})F_0(y_j),
\]
and
\[
b(j)=p_j^++p_j^-
=
1-F_0(y_j)
+
F_0(y_j/\sqrt{\mathrm{sf}})F_0(y_j).
\]
At \(y_j=y_*^{\mathrm{orig}}\), the balance equation \eqref{eq:balance_original} gives
\[
1-F_0(y_*^{\mathrm{orig}})
=
F_0(y_*^{\mathrm{orig}}/\sqrt{\mathrm{sf}})
F_0(y_*^{\mathrm{orig}}).
\]
Therefore,
\[
b(j_*)
=
2F_0(y_*^{\mathrm{orig}}/\sqrt{\mathrm{sf}})
F_0(y_*^{\mathrm{orig}}).
\]
Differentiating \(a(j)\) and evaluating at \(y_j=y_*^{\mathrm{orig}}\) gives
\[
-a'(j_*)
=
\log(\mathrm{sf})\,y_*^{\mathrm{orig}}
\left[
\phi(y_*^{\mathrm{orig}})
\left(
1+
F_0(y_*^{\mathrm{orig}}/\sqrt{\mathrm{sf}})
\right)
+
F_0(y_*^{\mathrm{orig}})
\frac{\phi(y_*^{\mathrm{orig}}/\sqrt{\mathrm{sf}})}
{\sqrt{\mathrm{sf}}}
\right].
\]
Substituting these expressions into \eqref{eq:fokker_planck_variance} yields
\[
\sigma_{J,*}^{2,\mathrm{orig}}
\approx
\frac{
F_0(y_*^{\mathrm{orig}}/\sqrt{\mathrm{sf}})
F_0(y_*^{\mathrm{orig}})
}
{
\log(\mathrm{sf})\,y_*^{\mathrm{orig}}
\left[
\phi(y_*^{\mathrm{orig}})
\left(
1+
F_0(y_*^{\mathrm{orig}}/\sqrt{\mathrm{sf}})
\right)
+
F_0(y_*^{\mathrm{orig}})
\dfrac{\phi(y_*^{\mathrm{orig}}/\sqrt{\mathrm{sf}})}
{\sqrt{\mathrm{sf}}}
\right]
}.
\]
Using \eqref{eq:balance_original} once more,
\(F_0(y_*^{\mathrm{orig}})[1+F_0(y_*^{\mathrm{orig}}/\sqrt{\mathrm{sf}})]=1\), and hence
\(F_0(y_*^{\mathrm{orig}}/\sqrt{\mathrm{sf}})F_0(y_*^{\mathrm{orig}})
=1-F_0(y_*^{\mathrm{orig}})\).
Multiplying numerator and denominator by \(F_0(y_*^{\mathrm{orig}})\) gives \eqref{eq:sigma_J_orig}.

For the modified update rule, \eqref{eq:transition_modified} gives
\(p_j^+=[1-F_0(y_j/\sqrt{\mathrm{sf}})][1-F_0(y_j)]\) and
\(p_j^-=F_0(y_j/\sqrt{\mathrm{sf}})F_0(y_j)\).
Hence
\[
a(j)=p_j^+-p_j^-
=
1-F_0(y_j/\sqrt{\mathrm{sf}})-F_0(y_j).
\]
At \(y_j=y_*^{\mathrm{mod}}\), \eqref{eq:balance_modified_symmetric} gives
\[
F_0(y_*^{\mathrm{mod}}/\sqrt{\mathrm{sf}})
+
F_0(y_*^{\mathrm{mod}})
=
1,
\]
so \(p_{j_*}^+=p_{j_*}^-=
F_0(y_*^{\mathrm{mod}}/\sqrt{\mathrm{sf}})F_0(y_*^{\mathrm{mod}})\), and therefore
\[
b(j_*)
=
2F_0(y_*^{\mathrm{mod}}/\sqrt{\mathrm{sf}})
F_0(y_*^{\mathrm{mod}}).
\]
Differentiating \(a(j)\) gives
\[
-a'(j_*)
=
\log(\mathrm{sf})\,y_*^{\mathrm{mod}}
\left[
\phi(y_*^{\mathrm{mod}})
+
\frac{\phi(y_*^{\mathrm{mod}}/\sqrt{\mathrm{sf}})}
{\sqrt{\mathrm{sf}}}
\right].
\]
Substituting this expression and \(b(j_*)\) into \eqref{eq:fokker_planck_variance} gives \eqref{eq:sigma_J_mod}.
This completes the proof.
\end{proof}
\stationaryspreadbscale*
\begin{proof}
We prove the statement for the original rule; the modified rule is identical.
By the delta-method conversion \eqref{eq:sigma_B_variants},
\[
\sigma_{B,*}^{\mathrm{orig}}
\approx
B_*^{\mathrm{orig}}\log(\mathrm{sf})\,\sigma_{J,*}^{\mathrm{orig}} .
\]
Using the equilibrium-size approximation \eqref{eq:B_star_variants}, we have
\[
B_*^{\mathrm{orig}}
\approx
\frac{v(1-\mathrm{sf}^{-1})}{S_\infty^2}
\frac{\left[y_*^{\mathrm{orig}}(\mathrm{sf})\right]^2}{\varepsilon^2}.
\]
Substitution gives
\[
\sigma_{B,*}^{\mathrm{orig}}
\approx
\frac{v(1-\mathrm{sf}^{-1})}{S_\infty^2}
\frac{
\left[y_*^{\mathrm{orig}}(\mathrm{sf})\right]^2
\log(\mathrm{sf})\,\sigma_{J,*}^{\mathrm{orig}}(\mathrm{sf})
}
{\varepsilon^2}.
\]
By Proposition~\ref{prop:stationary_variance_constants},
\(\sigma_{J,*}^{\mathrm{orig}}(\mathrm{sf})\) depends only on the scale factor and the update rule, and is independent of \(\varepsilon\).
Also, by Proposition~\ref{prop:equilibrium_scaling}, \(y_*^{\mathrm{orig}}(\mathrm{sf})\) is independent of \(\varepsilon\).
Therefore, for fixed \(\mathrm{sf}>1\),
\(\sigma_{B,*}^{\mathrm{orig}}=O(\varepsilon^{-2})\) as \(\varepsilon\downarrow0\).
The same substitution with \(B_*^{\mathrm{mod}}\) and
\(\sigma_{J,*}^{\mathrm{mod}}\) yields \eqref{eq:sigma_B_mod_scaling} and
\(\sigma_{B,*}^{\mathrm{mod}}=O(\varepsilon^{-2})\).

Squaring the two standard-deviation estimates gives
\[
(\sigma_{B,*}^{\mathrm{orig}})^2=O(\varepsilon^{-4}),
\qquad
(\sigma_{B,*}^{\mathrm{mod}})^2=O(\varepsilon^{-4}).
\]
This proves the proposition.
\end{proof}
\vskip 0.2in
\bibliography{refs, refs_stability}
\end{document}